\documentclass{article}

\PassOptionsToPackage{numbers,compress}{natbib}
\usepackage[main, final]{neurips_2026}

\usepackage[utf8]{inputenc} %
\usepackage[T1]{fontenc}    %
\usepackage[hidelinks]{hyperref}       %
\usepackage{url}            %
\usepackage{booktabs}       %
\usepackage{amsfonts}       %
\usepackage{nicefrac}       %
\usepackage{microtype}      %
\usepackage{xcolor}         %
\usepackage{enumitem}
\usepackage{amssymb}
\usepackage{graphicx}
\usepackage{subcaption}
\usepackage{orcidlink}

\graphicspath{{../docs/figures/}}

\usepackage{amsmath}
\usepackage{amsthm}

\usepackage{algorithm}
\usepackage{algpseudocode}

\theoremstyle{plain} %
\newtheorem{theorem}{Theorem}
\newtheorem{proposition}[theorem]{Proposition}
\newtheorem{lemma}[theorem]{Lemma}

\theoremstyle{definition} %
\newtheorem{definition}[theorem]{Definition}

\newcommand{\spara}[1]{\smallskip\noindent{\bf #1}}

\title{Overcoming the Resolution Limit: Significance-Aware Regularization for Intersectional Fairness}

\author{%
  Antonio Ferrara~\orcidlink{0000-0002-0784-1115} \quad Andr\'e Panisson~\orcidlink{0000-0002-3336-0374} \quad Francesco Cozzi~\orcidlink{0009-0001-5154-0651} \\
  \textbf{Alan Perotti~\orcidlink{0000-0002-1690-6865} \quad Francesco Bonchi~\orcidlink{0000-0001-9464-8315}} \\
  Intesa Sanpaolo AI Research, Turin, Italy \\
  \texttt{\{name.surname\}@intesasanpaolo.com} \\
}

\begin{document}

\maketitle

\begin{abstract}

Intersectional fairness requires evaluating models across fine-grained subgroups. However, as group definitions become more granular, the sample size per subgroup shrinks, causing fairness estimates to become increasingly unstable. Existing in-processing methods are \emph{certainty-blind}: they penalize observed disparities without distinguishing statistically certified unfairness from sampling noise.
This leads to \emph{fairness overfitting}, where models sacrifice utility to ``fix'' spurious disparities in sparse regimes. To address this, we introduce the  \emph{significant-excess} metric family, a continuous, uncertainty-aware fairness score that measures only the portion of a subgroup’s disparity exceeding what sampling variation can explain at a specific confidence level. Building on this metric, we propose SAFER (Significance-Aware Fairness Regularizer), a differentiable in-processing method that gates fairness penalties using a soft-count likelihood-ratio evidence statistic. SAFER satisfies a \emph{provable safety property}: when every subgroup disparity lies well below the gate threshold, the unconstrained optimum is nearly stationary for SAFER. At that optimum, the back-propagated contribution of its significance-gated component is exponentially small, and the additional fairness contribution comes from a small ungated term with a fixed weight. Mechanism ablations across three benchmark datasets and purpose-built stress variants show that removing any component of SAFER (threshold smoothing, the significance gate, or a dense-tail term) produces the failure the theory predicts on at least one stress variant. In sparse intersectional settings, SAFER achieves a better fairness--utility Pareto frontier than the baseline methods, and it preserves utility when certified unfairness is absent.

\end{abstract}

\section{Introduction}
\label{sec:introduction}

Intersectional fairness \cite{crenshaw,gohar2023} requires evaluating models across conjunctions of protected attributes, but as subgroups become more granular, their sample sizes inevitably shrink. In these sparse regimes, fairness estimates become noise-dominated, and intersectional fairness reaches a \emph{resolution limit}: the point at which the statistical uncertainty exceeds the disparity being measured \cite{Yang2021CausalIA,foulds2020intersectional,ferrara2025sizeadaptive}.
Indeed, even widely adopted heuristics such as the four-fifths rule lack statistical grounding and can both miss and fabricate disparities~\cite{raghavan2024}. Ignoring estimation uncertainty leads to \emph{fairness overfitting}, where the model memorizes group noise in the training set rather than genuine bias~\cite{laakom2025}. %
Existing in-processing methods are
\emph{certainty-blind}: they penalize observed disparities without
distinguishing certified unfairness from sampling noise, causing \emph{fairness
overfitting} in the sparse intersectional regime. Conversely, audit tools such as~\cite{ferrara2025sizeadaptive} incorporate uncertainty awareness, but only as 
a post-hoc binary accept/reject decision rather than a continuous metric or a differentiable training signal. If fairness regularization is to behave sensibly in sparse intersectional regimes, the optimization criterion itself must be able to distinguish well-supported disparities from noisy ones.

Whether gating on significance helps depends on how many training
subgroups lie above the gate threshold.  If none do, the gated fairness
contributions are suppressed.  If some do, the gate selects which
disparities receive greater weight.  If all do, the gate is effectively
inactive.

In this context, this paper makes three main contributions:
\begin{itemize}[leftmargin=*, noitemsep, topsep=1pt]
    \item \textbf{Uncertainty-aware fairness measurement
    (Section~\ref{sec:measurement}).} We introduce the \emph{significant-excess}
    metric family, which measures only the portion of each subgroup's disparity
    that exceeds what sampling variation can explain at a given confidence level. The construction applies to any parity criterion expressible
    as a continuously differentiable function of multinomial cell probabilities, including
    Statistical Parity \cite{dwork2012}, Disparate
    Impact \cite{feldman2015certifying}, Equalized Odds and Equal Opportunity \cite{hardt2016equality}, and yields a continuous, uncertainty-aware fairness score that is zero
    when no violation is statistically certified.

\smallskip

    \item \textbf{Significance-gated optimization
    (Section~\ref{sec:safer}).} We propose \textbf{SAFER} (Significance-Aware
    Fairness Regularizer), a differentiable in-processing method that gates
    fairness penalties using a soft-count likelihood-ratio evidence
    statistic.  The gate attenuates subgroup penalties according to their
    evidence of disparity; statistical certification comes from the
    held-out audit. SAFER satisfies a provable \emph{safety
    property}: when every disparity lies well below the gate
    threshold, the optimum of unconstrained empirical risk
    minimization~(ERM) is nearly stationary for SAFER. At that optimum, the back-propagated
    contribution of its significance-gated component is exponentially small, and the
    additional fairness contribution comes from a small ungated term with a fixed
    weight.

\smallskip

    \item \textbf{Regime characterization and empirical validation
    (Section~\ref{sec:experiments} and Appendix~\ref{app:synthetic}).} We identify three operating
    regimes (safety, selective gating, gate always open), defined by
    which training subgroups cross the gate's threshold, and test each
    through mechanism ablations on purpose-built stress variants and a
    controlled benchmark.
\end{itemize}

The claims are regime-conditional: SAFER gives a better Pareto
frontier than the baselines where certified disparities exist and preserves the utility of
the unconstrained model where they do not, and removing any component of SAFER
produces the failure the theory predicts on at least one stress variant.

\section{Related Work}
\label{sec:related-work}

\spara{Intersectional and subgroup fairness.}
Equality of Opportunity~\cite{hardt2016equality} and related parity criteria
are the standard building blocks for group fairness, but models can satisfy
coarse constraints yet violate fairness on structured
intersections~\cite{kearns2018preventing}. Differential fairness treats
intersectionality as a first-class object~\cite{islam2023differential};
worst-case comparisons formalize the combinatorial structure of multi-attribute
subgroups~\cite{ghosh2021characterizing}; and
multicalibration~\cite{hebertjohnson2018multicalibration} provides simultaneous
calibration guarantees over rich subgroup families. 
Molina and Loiseau~\cite{Molina2022} bound intersectional violations via marginal constraints; Himmelreich et al.~\cite{Himmelreich2025} formalize the tension between group granularity and statistical power. %
Our work targets the same
setting but a different bottleneck: once many intersectional groups are audited
simultaneously, their disparities become heterogeneous in statistical
reliability, and this reliability must enter both evaluation and optimization.

\spara{Uncertainty-aware fairness.}
SAFT~\cite{ferrara2025sizeadaptive} is the work closest to ours: it
tests whether each subgroup's confidence interval excludes parity, applying
multiplicity corrections across the subgroup family. This makes subgroup size
and estimator variance part of the fairness decision; certainty-blind metrics~\cite{hort2024bias} ignore both. We build on SAFT in two
directions. First, we replace binary per-subgroup flags with a continuous
\emph{significant-excess} score that preserves severity information beyond the
accept/reject decision, yielding a single metric for ranking models
(Section~\ref{sec:measurement}). Second, this statistical logic
motivates a differentiable evidence gate that modulates subgroup
penalties during training.
SAFER keeps SAFT's intervals for the audit and trains with a
soft-count likelihood-ratio statistic, because a plug-in Wald standard
error is unstable on small soft counts and a posterior-sampling test is
too expensive inside a training loop (Section~\ref{sec:safer-lrt}).
More broadly, a growing body of work grounds fairness evaluation in statistical inference: Cherian and Candes~\cite{Cherian2024} provide finite-sample valid tests for auditing; Tang et al.~\cite{tang2026} replace Wald statistics with empirical-likelihood tests; and Lo et al.~\cite{lo2024} advocate classical hypothesis testing for AI fairness. On the evaluation side, Barrainkua et al.~\cite{barrainkua2024} show that fairness conclusions are unstable under resampling, and Rosenblatt and Witter~\cite{rosenblatt2024} benchmark how uncertainty can reverse fairness rankings. Kuzucu et al.~\cite{Kuzucu2025} treat predictive uncertainty itself as a fairness criterion, whereas we use statistical uncertainty of the fairness \emph{estimate} to gate optimization.
Besse et al.~\cite{besse2018} were among the first to propose
  confidence intervals for testing disparate impact, which replaced
  point estimates with interval-based decisions.

\spara{Fairness-aware optimization.}
A large literature mitigates unfairness during or after training.
Post-processing methods adjust a fixed classifier's
thresholds~\cite{hardt2016equality}, but cannot reshape the learned
representation and become impractical when the number of intersectional
subgroups is combinatorially large. Among in-processing approaches,
Exponentiated Gradient~\cite{agarwal2018reductions} solves a sequence of
weighted learning problems, Adversarial
Debiasing~\cite{zhang2018mitigating} and Prejudice
Remover~\cite{kamishima2012fairness} embed fairness pressure through an
adversary or a dependence penalty, and differentiable regularizers such as
Bechavod and Ligett~\cite{bechavod2017penalizing} and
\textsc{fairret}~\cite{buyl2024fairret} penalize group discrepancy via smooth
surrogates. 
Recent intersectional-specific optimizers include Lee et al.~\cite{lee2025}, who regularize over multitype attributes, and N\v{e}me\v{c}ek et al.~\cite{nemecek2026}, who enforce exact intersectional constraints via mixed-integer programming. Both remain certainty-blind. Distributionally robust optimization (DRO)~\cite{hashimoto2018fairness} and minimax group
fairness~\cite{diana2021minimax} minimize worst-case subgroup loss, while
FairDRO~\cite{park2025fairdro} casts group fairness as classwise robust
optimization. All these methods are \emph{certainty-blind}: once a subgroup disparity is observed, it
contributes to the objective regardless of whether it is statistically
distinguishable from sampling noise. SAFER differs by using a
likelihood-ratio evidence gate to modulate disparity penalties during training,
while retaining the held-out audit for statistical certification.

\section{Measuring Intersectional Fairness}
\label{sec:measurement}

We introduce the \textit{significant-excess metric} family to address the \emph{resolution limit} identified in Section~\ref{sec:introduction}. To this end, these metrics are designed to answer two questions simultaneously: does a
subgroup exhibit a non-zero disparity compared to the rest of the population, and is that disparity statistically significant enough to be
distinguished from sampling noise?

\spara{Setup.}
\label{sec:measurement-setup}
Let $\mathfrak{S}$ be a family of protected subgroups, including
intersectional groups formed by conjunctions of sensitive attributes. For each
subgroup $S \in \mathfrak{S}$, we compare the behavior of a model on $S$ and its
complement $\bar{S}$. Both are sets of individuals: $S$ collects those
with the subgroup's attribute values and $\bar S$ the rest of the
population. We write $\phi(S)$ for a generic fairness functional expressible as a continuously differentiable function of multinomial cell probabilities. We write $\phi_n(S)$ for its empirical estimate on a test sample of size $n$. For each subgroup disparity $\phi_n(S)$ we require a confidence interval of level $(1{-}\alpha)$, expressible by a lower and upper bound $[\ell_\alpha(S),\, u_\alpha(S)]$, which quantifies the uncertainty of the estimation. 
We assume $\ell_\alpha(S)\le\phi_n(S)\le u_\alpha(S)$.

\begin{definition}[Significant excess]
\label{def:significant-excess}
For a subgroup $S$ and significance level $\alpha$, we define the
\emph{significant excess} of fairness criterion $\phi$ as
\vspace{-10pt}
\begin{equation}
\label{eq:significant-excess}
    \mathcal{E}^{\phi}_{\alpha}(S)
    \;=\;
    \begin{cases}
        \max \bigl(0,\; \ell_\alpha(S)\bigr)
            & \text{if } \phi_n(S) \geq 0, \\[4pt]
        \max \bigl(0,\; -u_\alpha(S)\bigr)
            & \text{if } \phi_n(S) < 0.
    \end{cases}
\end{equation}
\end{definition}

The significant excess is the distance from zero to the interval.
When the interval contains zero, the disparity is not
statistically certified and the contribution is exactly zero. When the interval
excludes zero, $\mathcal{E}^{\phi}_{\alpha}(S)$ measures the portion of the observed
disparity that remains after subtracting what sampling noise can plausibly
explain.

\spara{Basic properties.}
\label{prop:significant-excess-properties}
The significant-excess score satisfies:
\begin{enumerate}[label=\textbf{\arabic*.}, leftmargin=*, noitemsep, topsep=1pt]
\item \textbf{Zero iff no certified unfairness under the interval rule.}
    $\mathcal{E}^{\phi}_{\alpha}(S)=0$ if and only if the interval for $\phi(S)$
    contains zero.
\item \textbf{Scale preservation.}
    $\mathcal{E}^{\phi}_{\alpha}(S)$ is expressed in the same units as $\phi$.
\item \textbf{Uncertainty awareness.}
    The score is nonincreasing when the interval expands by inclusion.
\item \textbf{Continuity.}
    The score is continuous in the interval endpoints.
\end{enumerate}

Finally, we refer to the \textit{Mean Significant Measure Excess} as the mean aggregation over the nonempty audited family of subgroups with defined estimates and intervals:
\vspace{-5pt}
\begin{equation}
\label{eq:aggregate-excess}
    \overline{\mathcal{E}}^{\phi}_{\alpha}(f)
    =
    \frac{1}{|\mathfrak{S}|}
    \sum_{S \in \mathfrak{S}} \mathcal{E}^{\phi}_{\alpha}(S).
\end{equation}

Let us consider two fairness metrics as an example, the Statistical Parity (SP) gap and the Equal Opportunity (EO) gap:
\begin{align}
    SP(S) &= P(\hat{Y}=1 \mid S) - P(\hat{Y}=1 \mid \bar{S}), \\
    EO(S) &= P(\hat{Y}=1 \mid Y=1, S) - P(\hat{Y}=1 \mid Y=1, \bar{S}).
\end{align}

For SP, the relevant cells are predicted
positive/negative outcomes conditioned on $S$ and $\bar{S}$. For EO, the same construction
is applied after conditioning on the positive-label population $Y=1$.
Both can be written as continuously differentiable functions of
multinomial cell probabilities. Given a confidence level $\alpha$, we can define the \textit{Mean Significant Parity Excess} (MSPE) by applying Equations \ref{eq:significant-excess} and \ref{eq:aggregate-excess} with $\phi(S) = SP(S)$, and the \textit{Mean Significant Equal-Opportunity Excess} (MSEE) with $\phi(S) = EO(S)$.
\begin{equation}
\label{eq:mspe-msee}
    MSPE_{\alpha}(f) \;=\; \overline{\mathcal{E}}^{SP}_{\alpha}(f),
    \qquad
    MSEE_{\alpha}(f) \;=\; \overline{\mathcal{E}}^{EO}_{\alpha}(f).
\end{equation}
We obtain the confidence intervals $[\ell_\alpha(S),\, u_\alpha(S)]$ with the SAFT framework~\cite{ferrara2025sizeadaptive}, because its test adapts to both large and small subgroups.

\section{SAFER: A Differentiable Surrogate for Significant-Excess Metrics}
\label{sec:safer}

$MSEE_\alpha$ (Section~\ref{sec:measurement}) measures only the
statistically certified portion of each subgroup's disparity.  It is not
directly trainable, because it depends on hard predictions, which
are piecewise constant in~$\theta$, and on an absolute value and a
positive part, which are not differentiable at zero.  SAFER
replaces these with differentiable approximations and is built around two
ideas.  First, a \emph{soft-count likelihood-ratio test} (LRT) provides a
differentiable significance statistic that requires no explicit variance
estimation.  This matters in the sparse intersectional cells,
where Wald-type statistics become unreliable (\S\ref{sec:safer-lrt}).  Second,
a significance gate built from this statistic yields a provable
\emph{safety property}: when every subgroup disparity lies well
below the gate threshold, the ERM optimum is nearly stationary for SAFER. At that optimum, the back-propagated
contribution of the significance-gated component is exponentially small, and the
additional fairness contribution is proportional
to the weight of a small ungated term
(\S\ref{sec:safer-safety}).  Proofs are in
Appendix~\ref{app:proofs-v2}.

\subsection{Threshold-Smoothed Soft Counts}
\label{sec:safer-thresholding}

Let $f_\theta:\mathcal X\to[0,1]$ be the model score.  For each
subgroup $S\in\mathfrak S$ write $\Pi_S$ for the conditioning
population: $\Pi_S=S$ for Statistical Parity, and
$\Pi_S=(S,\,Y{=}1)$ for Equal Opportunity, matching
Section~\ref{sec:measurement}.  Define three population positive rates, relative to the hard binary predictions, the raw predicted probabilities and the $\tau$ smoothed sigmoid:
\begin{align}
\label{eq:phi-hard}
    p^{\mathrm{hard}}_S(\theta)
    &= \mathbb{E}\!\left[
       \mathbf{1}\{f_\theta(X)>\tfrac12\}
       \;\middle|\; \Pi_S\right],
    \\
\label{eq:phi-raw}
    p^{\mathrm{raw}}_S(\theta)
    &= \mathbb{E}\!\left[
       f_\theta(X)
       \;\middle|\; \Pi_S\right],
    \\
\label{eq:phi-tau}
    p^\tau_S(\theta)
    &= \mathbb{E}\!\left[
       \sigma_\tau\!\left(f_\theta(X)-\tfrac12\right)
       \;\middle|\; \Pi_S\right],
    \qquad
    \sigma_\tau(u)=\sigma(u/\tau),
\end{align}
where $\sigma$ is the logistic sigmoid.  The corresponding disparity is
$\phi^\bullet_S = p^\bullet_S - p^\bullet_{\bar S}$ for each variant
$\bullet\in\{\mathrm{hard},\mathrm{raw},\tau\}$. For example in this notation, $\phi^{\mathrm{hard}}_S$ is exactly the
population functional $\phi(S)$ of Section~\ref{sec:measurement}. Furthermore, we use hats
(e.g., $\widehat\phi^\tau_S$) to denote the empirical analogues computed from sample soft counts.
The decision threshold $1/2$ is the Bayes rule when the scores
estimate class probabilities and both error types cost the same.  It is
a property of the deployed classifier and is fixed before training.  The
construction applies unchanged to any other fixed threshold $t$ by
replacing $f_\theta(x)-\tfrac12$ with $f_\theta(x)-t$;
Appendix~\ref{app:threshold-sweep} reports SAFER retrained and audited
at thresholds from $0.1$ to $0.9$.

In particular, replacing $\mathbf{1}\{f_\theta(x)>\frac12\}$ with
$\sigma_\tau(f_\theta(x)-\frac12)$ yields a differentiable
approximation whose quality is controlled linearly by~$\tau$:

\begin{theorem}
\label{thm:thresh-consistency}
If, for some $\delta>0$, the conditional score distributions on $\Pi_S$ and $\Pi_{\bar S}$
admit densities bounded by~$L$ on $\{u:|u-\tfrac12|\le\delta\}$, then
$|\phi^\tau_S(\theta)-\phi^{\mathrm{hard}}_S(\theta)|\le 4L\tau\log 2{}+2e^{-\delta/\tau}$.
\end{theorem}

A natural alternative, using the raw probability $f_\theta(x)$
directly ($p^{\mathrm{raw}}_S$), does not work:

\begin{proposition}[Raw soft counts are not hard-decision consistent]
\label{prop:raw-inconsistency}
For every $\eta>0$, there exist continuous bounded-density score
distributions such that
$|\phi^{\mathrm{raw}}_S(\theta)-\phi^{\mathrm{hard}}_S(\theta)|
\ge1-\eta$,
and no constant rescaling of $\phi^{\mathrm{raw}}_S$ matches
$\phi^{\mathrm{hard}}_S$ uniformly.
\end{proposition}
Two groups sitting just across the $\frac12$ boundary from opposite
sides produce a hard gap near~$1$ but a raw-probability gap near~$0$;
two groups both well above the boundary produce the reverse.  No single
correction can fix both regimes, so raw soft counts can rank models
differently from the hard-decision audit.  Threshold smoothing also
localizes gradients to examples near the decision boundary, where
predictions can flip (Theorem~\ref{thm:gradient-localization}
in Appendix~\ref{app:proofs-v2}).

\subsection{A Differentiable LRT for Subgroup Disparities}
\label{sec:safer-lrt}
\label{sec:safer-direct}

Given threshold-smoothed soft counts, we need a differentiable
evidence statistic to select subgroup disparities for training.
An approach would be the Bernoulli-form Wald statistic
$|Z_S|=|\widehat\phi^\tau_S|/\widehat{SE}^\phi_S$, where $\widehat{SE}^\phi_S$ is the two-proportion standard-error proxy of property (ii), but this has two
problems in the sparse intersectional regime.  First,
$\widehat{SE}^\phi_S$ is numerically unstable when soft counts are
small.  The second problem matters more for optimization: the Wald loss
can decrease by \emph{inflating the standard-error estimate} rather than
reducing the disparity, since both numerator and denominator depend
on~$\theta$.

SAFER instead uses the Bernoulli-form two-sample deviance
statistic computed on threshold-smoothed soft counts.  For subgroup~$S$ define the soft counts
$
    \tilde n_{1,S}(\theta)
    = \sum_{i\in\Pi_S}\sigma_\tau\!\bigl(f_\theta(x_i)-\tfrac12\bigr)$
and $
    \tilde n_{0,S}(\theta)
    = |\Pi_S| - \tilde n_{1,S}(\theta)$.  Let $\tilde n_S=\tilde n_{1,S}+\tilde n_{0,S}$ and define analogous
quantities for~$\bar S$.  Write
$\hat p_S=\tilde n_{1,S}/\tilde n_S$,
$\hat p_{\bar S}=\tilde n_{1,\bar S}/\tilde n_{\bar S}$, and
$\hat p=(\tilde n_{1,S}+\tilde n_{1,\bar S})/
(\tilde n_S+\tilde n_{\bar S})$ for the pooled null estimate.  The likelihood-ratio
statistic for the soft-parity null $H_0\colon p^\tau_S=p^\tau_{\bar S}$ is
\begin{equation}
\label{eq:lrt-soft}
    \Lambda_S(\theta)
    =
    -2\!\left[
        \tilde n_{1,S}\log\frac{\hat p}{\hat p_S}
        + \tilde n_{0,S}\log\frac{1{-}\hat p}{1{-}\hat p_S}
        + \tilde n_{1,\bar S}\log\frac{\hat p}{\hat p_{\bar S}}
        + \tilde n_{0,\bar S}\log\frac{1{-}\hat p}{1{-}\hat p_{\bar S}}
    \right].
\end{equation}
For SP, these are soft counts of predicted positive/negative across~$S$
and~$\bar S$; for EO, the same construction restricts to the
positive-label population.

Three properties make the soft-count LRT a suitable significance
statistic for intersectional fairness training:
\begin{enumerate}[label=\textbf{(\roman*)},leftmargin=*,noitemsep,topsep=2pt]
\item \textbf{No variance estimation.}
    $\Lambda_S$ is a closed-form function of the soft
    counts: it never estimates or differentiates through
    $\widehat{SE}^\phi_S$.  This removes the gradient path through which
    the Wald loss can be lowered by inflating the standard error.
\item \textbf{Boundary stability.}
    In sparse cells the Bernoulli-form variance proxy
    $(\widehat{SE}^\phi_S)^2
      = \hat p_S(1-\hat p_S)/\tilde n_S
      + \hat p_{\bar S}(1-\hat p_{\bar S})/\tilde n_{\bar S}$
    is unstable: with only a few effective observations,
    threshold-smoothed rates can approach zero or one, the variance
    estimate collapses toward zero, and $|Z_S|$ grows without bound,
    so a few tiny, saturated subgroups can dominate the objective.
    $\Lambda_S$ is computed from the Bernoulli log-likelihood itself and
    contains no division by a plug-in variance estimate that can vanish,
    so it avoids this boundary singularity.
\item \textbf{Asymptotic equivalence.}
    For interior rates and vanishing disparities, a second-order
    expansion gives $\Lambda_S=Z_S^2\,(1+o_p(1))$, so the two
    statistics use the same reference cutoff $z_\alpha$.  For a fixed
    predictor evaluated on independent observations under soft parity,
    $\Lambda_S\Rightarrow c\chi^2_1$, $0\le c\le1$, and the Bernoulli
    reference cutoff is asymptotically conservative
    (Appendix~\ref{app:property-iii}).
\end{enumerate}
During training the gate is an optimization device; statistical
certification is provided by the held-out hard-count audit of
Section~\ref{sec:measurement}.

\spara{LRT-based threshold scale.}
To express the significance threshold in the raw-disparity units of
$MSEE_\alpha$ when the disparity is nonzero, we define an effective
standard-error proxy from the empirical disparity and the deviance:
\begin{equation}
\label{eq:se-eff}
    \widehat{SE}_{\mathrm{eff},S}(\theta)
    =
    \frac{|\widehat\phi^\tau_S(\theta)|}{\sqrt{\Lambda_S(\theta)}}\,.
\end{equation}
This conversion puts the dimensionless evidence statistic on the
disparity scale, so that the training threshold carries the units of
$\phi$.  Under the local conditions of property (iii),
$\widehat{SE}_{\mathrm{eff},S}=\widehat{SE}^\phi_S(1+o_p(1))$,
recovering the Bernoulli-form normalization.  Before numerical clamping,
the threshold therefore scales with $\sqrt{1/n_S+1/n_{\bar S}}$ for SP
and $\sqrt{1/m_S+1/m_{\bar S}}$ for EO.  It thus shrinks as the inverse
square root of the sample size when both comparison groups grow in
proportion.
The implementation
regularizes the ratio and clamps it for numerical stability.

\spara{The trained objective.}
SAFER penalizes excess disparity above the LRT-based training
threshold, with the threshold detached from the gradient
computation:
\begin{equation}
\label{eq:direct-surrogate}
    \mathcal L_{\mathrm{dir}}(\theta)
    =
    \frac{1}{|\mathfrak S|}
    \sum_{S\in\mathfrak S}
    \operatorname{softplus}_{\tau_s}\!\left(
        \sqrt{\widehat\phi^\tau_S(\theta)^2+\varepsilon^2}
        \;-\;
        z_\alpha\;\operatorname{sg}\!\left[
        \widehat{SE}_{\mathrm{eff},S}(\theta)\right]
    \right),
\end{equation}
where $\operatorname{softplus}_\tau(u)=\tau\log(1+e^{u/\tau})$ and
$\operatorname{sg}[\cdot]$ denotes stop-gradient (the enclosed
expression is treated as a constant during back-propagation).
Detachment ensures that the optimizer can only reduce the loss by
shrinking $\widehat\phi^\tau_S$, not by inflating the threshold.  The
threshold still updates between optimizer steps as $\theta$ changes, so
it remains adaptive: it does not provide a gradient path.
Within one update the gated term therefore gives the optimizer no
first-order incentive to raise the threshold; across updates the
threshold follows the predictions, which every term of the objective
changes.  Appendix~\ref{app:stop-gradient} gives a scalar example and a
diagnostic over training runs, in which the mean threshold shows no
upward drift.  A canonical version of Eq.~\eqref{eq:direct-surrogate},
in which the surrogate and the metric share the population Wald threshold $B_S=z_\alpha SE^\phi_S$,
approximates $MSEE^{\mathrm{pop}}_\alpha$ (Eq.~\eqref{eq:canonical-msee})
pointwise with error at most
$\varepsilon+\tau_s\log 2+4L\tau\log 2{}+2e^{-\delta/\tau}$ (Theorem~\ref{thm:surrogate-approx}
in Appendix~\ref{app:proofs-v2}).

\subsection{Safety Property and Operating Regimes}
\label{sec:safer-safety}
\label{sec:safer-densetail}

The softplus gate in $\mathcal L_{\mathrm{dir}}$ separates
subgroups above the gate threshold from those below it.  Let
\[
    u_S(\theta)
    =
    \sqrt{\widehat\phi^\tau_S(\theta)^2+\varepsilon^2}
    \;-\;
    z_\alpha\;\operatorname{sg}\!\left[
    \widehat{SE}_{\mathrm{eff},S}(\theta)\right].
\]
Since the stop-gradient ($\operatorname{sg}$) freezes the threshold for differentiation,
$\nabla_\theta u_S =
\nabla_\theta\sqrt{\widehat\phi^\tau_S(\theta)^2+\varepsilon^2}$.
Throughout, $\nabla_\theta$ of an expression containing
$\operatorname{sg}$ denotes the back-propagated fairness contribution
with the detached quantity held fixed;
Proposition~\ref{cor:safety} bounds it.
The chain rule and
$\operatorname{softplus}'_{\tau_s}(u)=\sigma(u/\tau_s)$ give
\begin{equation}
\label{eq:hinge-suppression}
    \bigl\|\nabla_\theta
    \operatorname{softplus}_{\tau_s}(u_S)\bigr\|
    \le
    \sigma(u_S/\tau_s)\;
    \bigl\|\nabla_\theta u_S\bigr\|.
\end{equation}
When $u_S<0$ (disparity below the gate threshold), the
prefactor $\sigma(u_S/\tau_s)$ decays exponentially as
$e^{u_S/\tau_s}$, suppressing the fairness gradient on that subgroup.

\spara{Dense tail.}
Under strong regularization, active disparities may stay
close to the hinge boundary where the softplus gate attenuates the gradient
substantially: by Eq.~\eqref{eq:hinge-suppression} the gradient is
multiplied by $\sigma(u_S/\tau_s)$, which equals $1/2$ at the
gate boundary and is already below $0.01$ at $u_S=-0.05$ for
$\tau_s=10^{-2}$, so under strong regularization the optimization can
stall before an active disparity is removed.
To prevent stalling, SAFER adds a small ungated term:
\vspace{-5pt}
\begin{equation}
\label{eq:dense-tail}
    \mathcal L_{\mathrm{tail}}(\theta)
    =
    \frac{w_{\mathrm{tail}}}{|\mathfrak S|}
    \sum_{S\in\mathfrak S}
    \sqrt{\Lambda_S(\theta)}\,,
\end{equation}
where $w_{\mathrm{tail}}\ge0$.  This term carries no
gate multiplier, so it provides a gradient even when the
softplus is saturated.  Since $\sqrt{\Lambda_S}$ locally expresses the disparity in units of the
Bernoulli-form normalization, the tail weights each subgroup by its statistical
support (Appendix~\ref{app:dense-tail}).  Because the term is ungated,
the safety bound below includes an additional term proportional to
$w_{\mathrm{tail}}$, while the gated part of the gradient keeps its
exponential bound; with $w_{\mathrm{tail}}=0$ the objective is gated
only.  We use the fixed value $w_{\mathrm{tail}}=0.003$ in all
runs of the full method.
When active subgroups dominate, the tail is negligible relative to
$\mathcal L_{\mathrm{dir}}$.

\begin{proposition}[Safety at the unconstrained optimum]
\label{cor:safety}
If $u_S(\theta^\star_{\mathrm{ERM}})\le-\epsilon^\star$ for every
$S\in\mathfrak S$, and the score and statistic gradients are bounded
near $\theta^\star_{\mathrm{ERM}}$, then
\begin{align}
\label{eq:safety-bound}
    \bigl\|\nabla_\theta\mathcal L_{\mathrm{dir}}
    (\theta^\star_{\mathrm{ERM}})\bigr\|
    &= O\!\left(e^{-\epsilon^\star/\tau_s}\right),
    \\
\label{eq:safety-tail-bound}
    \bigl\|\nabla_\theta
    (\mathcal L_{\mathrm{dir}}+\mathcal L_{\mathrm{tail}})
    (\theta^\star_{\mathrm{ERM}})\bigr\|
    &\le
    O\!\left(e^{-\epsilon^\star/\tau_s}\right)
    +
    \frac{w_{\mathrm{tail}}}{|\mathfrak S|}
    \sum_{S\in\mathfrak S}
    \bigl\|\nabla_\theta
    \sqrt{\Lambda_S(\theta^\star_{\mathrm{ERM}})}\bigr\|.
\end{align}
\vspace{-15pt}
\end{proposition}
When every subgroup lies below the gate boundary by a
margin $\epsilon^\star$ large relative to $\tau_s$, the ERM optimum is
nearly stationary for SAFER: the gated
fairness contribution there is exponentially
small, and the dense-tail residual (Eq.~\eqref{eq:dense-tail}) enters
the full objective multiplied by $\lambda w_{\mathrm{tail}}$.

\spara{Three operating regimes.}
The gate threshold determines which subgroups contribute to the
fairness gradient.  Write
$\mathcal A(\theta)=\{S\in\mathfrak S:u_S(\theta)>0\}$ for the
\emph{active set} of training subgroups: a subgroup is active
when $u_S>0$, which, up to the smoothing constant $\varepsilon$ and the
numerical clamps, means $T_S=\sqrt{\Lambda_S}>z_\alpha$, with
$z_\alpha=1.96$ at $\alpha=0.05$.  At a given checkpoint $\theta$, the
active set places a dataset in one of three regimes:
\begin{enumerate}[label=\textbf{(\roman*)},leftmargin=*,noitemsep,topsep=0pt]
\item \textbf{Safety} ($\mathcal A(\theta)=\varnothing$): when all
    subgroups at the ERM point lie well below the boundary relative to
    $\tau_s$, the gated fairness
    contribution is exponentially small, and the unconstrained optimum is
    nearly stationary for SAFER up to the dense-tail residual.
\item \textbf{Selective gating} ($0<|\mathcal A(\theta)|<|\mathfrak S|$,
    active and inactive subgroups coexist): the gate
    concentrates model capacity on the disparities for which the data
    provide statistical support.
\item \textbf{Gate always open} ($\mathcal A(\theta)=\mathfrak S$, all
    subgroups above the boundary): when all lie well above it relative to
    $\tau_s$, the gate is effectively inactive, and
    SAFER approaches a certainty-blind penalty.
\end{enumerate}
The regime is read off the active set at a checkpoint; when we
characterize a dataset we use the checkpoint of the unconstrained model,
and the regime can change during training.  Whether a subgroup is
active depends on its disparity, on its outcome rates, and on the
counts on both sides of the comparison (Appendix~\ref{app:regimes}), so
the boundary between regimes is set by the evidence in each subgroup.
Appendix~\ref{app:synthetic} states the hypothesis these regimes
suggest and tests it on a controlled benchmark.

\spara{Full objective.}
\label{sec:safer-objective}
\label{sec:safer-beyond}
For fairness criterion $\phi$, SAFER minimizes
\begin{equation}
\label{eq:safer-objective}
    \mathcal L_{\mathrm{SAFER}}(\theta)
    =
    \mathcal L_{\mathrm{pred}}(\theta)
    \;+\;
    \lambda\bigl[
        \mathcal L_{\mathrm{dir}}(\theta)
        + \mathcal L_{\mathrm{tail}}(\theta)
    \bigr],
\end{equation}
where $\mathcal L_{\mathrm{pred}}$ is binary cross-entropy and
$\lambda\ge0$ controls the fairness--utility tradeoff.

\spara{Extensions.}
The construction extends to any fairness functional differentiable in
multinomial cell probabilities: substituting SP gives MSPE directly,
Equalized Odds adds an FPR-parity hinge, and Disparate Impact uses the
log-DI metric with the same SP parity null
(Appendix~\ref{app:beyond-msee}).

\section{Experiments}
\label{sec:experiments}

The experiments serve two purposes.  First, we compare SAFER on the
Pareto frontier against the strongest available in-processing fairness
baselines on three datasets spanning sample sizes and disparity regimes.
Second, we use constructed \emph{stress variants} of the datasets
(low training budgets, deeper intersection levels and small validation
budgets) to probe the three mechanisms predicted by Section~\ref{sec:safer}.
Each variant targets a setting in which one SAFER component should
be needed, so removing that component in an ablation should have a
measurable effect. The full codebase is available at \url{https://github.com/Ambress92/Safer}.

\spara{Datasets and Stress Variants.}
\label{sec:exp-datasets}
Public benchmarks with multiple protected attributes and non-trivial
intersectional structures are scarce.  The main figures use the three
most established ones (Adult Income~\cite{kohavi1996scaling},
COMPAS~\cite{angwin2022machine} and German
Credit~\cite{statlog_german_credit_data_144}).
Table~\ref{tab:datasets} summarizes their stress variants, each of which
targets one of the mechanisms of Section~\ref{sec:safer}.  The table also reports ACS
Coverage~\cite{ding2021retiring} and Taiwan
Credit~\cite{yeh2009comparisons}, which add a public-coverage task with larger samples and a
credit-risk task with many intersections.  Unless a train-test split or intersection level is stated explicitly, dataset
names below refer to the default variants in Table~\ref{tab:datasets}.  The
default COMPAS variant uses the more severe 20/80 train-test split and is reserved for
mechanism ablations; the headline COMPAS comparison instead uses a balanced
50/50 split for a reasonable training budget.

\begin{table}[t]
\centering
\small
\caption{Datasets, stress variants, and the mechanism each variant
probes.  $|\mathfrak{S}|$ is the number of audited intersectional
subgroups at the indicated level; $\bar n_S$ is the mean training-set
subgroup size; $\bar m_S$ is the mean positive-label subgroup size
relevant to EO.  All values are approximate averages over splits.}
\label{tab:datasets}
\resizebox{\linewidth}{!}{%
\begin{tabular}{lcccccc}
\toprule
Dataset & Level & Split & $n_{\mathrm{train}}$ & $|\mathfrak{S}|$ &
$\bar n_S \;/\; \bar m_S$ & Targeted mechanism \\
\midrule
Adult Income     & 4 & 10/90 & 4,522         & 228 &
19.9 / 4.9  & threshold smoothing: high intersectionality, low budget \\
COMPAS           & 3 & 20/80 & 993           & 35  &
28.4 / 14.2 & dense tail: small training set, strong regularization \\
German Credit    & 3 & 50/50 & 500           & 32  &
15.6 / 10.9 & significance gate: safety regime \\
ACS Coverage     & 3 & 50/50 & 25,000        & 71  &
352.1 / 103.5 & higher-sample ACS benchmark \\
Taiwan Credit    & 4 & 50/50 & 15,000        & 285 &
52.6 / 11.6 & high-intersectionality credit risk \\
\bottomrule
\end{tabular}}
\vspace{-5pt}
\end{table}

Adult probes the threshold-smoothing mechanism
(almost every intersection has few samples). COMPAS
probes the dense-tail mechanism under strong regularization (small training set, moderate
disparity). German probes the safety
regime (almost no SP/EO violation is statistically certifiable at
\(\alpha=0.05\) under ERM in our observed splits).  ACS Coverage has larger effective subgroup sizes,
while Taiwan Credit tests whether the comparison extends to a larger
credit-risk task with many audited intersections but a relatively small
positive-label budget.  Appendix~\ref{app:datasets} details preprocessing,
protected attributes, subgroup construction, and split-level audit statistics.

\spara{Protocol and Metrics}
\label{sec:exp-protocol}
Each method and \(\lambda\) is evaluated over at least 10 stratified shuffle
splits, with at least 3 random initialization seeds per split
(Appendix~\ref{app:training} gives the counts per experiment).  All headline neural
methods share the same MLP backbone, prediction-loss schedule, and
held-out validation-only model selection.
We report \(MSEE_\alpha\)
at \(\alpha=0.05\) on hard test predictions
(Eq.~\eqref{eq:mspe-msee}) against held-out log-loss, with standard error (SE) bars
over the runs across splits and seeds.
MSEE and MSPE use unadjusted per-subgroup intervals; reported violation
counts use Benjamini--Hochberg (BH)-corrected subgroup $p$-values.
Hyperparameter grids, BCE-guarded
checkpointing, and clustered-SE checks are in
Appendix~\ref{app:training}.

\subsection{Pareto Comparison Against Baselines}
\label{sec:exp-pareto}

We compare SAFER against ERM, EO-Penalty and two
FAIRRET~\cite{buyl2024fairret} regularizers.  EO-Penalty is the
certainty-blind squared-disparity regularizer
\((\mathcal L_{\rm pred}+\frac{\lambda}{|\mathfrak S|}\sum_S
(\widehat\phi^{\mathrm{raw}}_S)^2)\) and is the closest direct
comparator to the SAFER significant-excess hinge.  FAIRRET Norm 1 and
FAIRRET Norm 2 are ratio-normalized differentiable fairness
regularizers.  EO-Penalty and
both FAIRRET variants operate on raw soft
probabilities rather than threshold-smoothed predictions, which makes
them direct empirical tests of the inconsistency claim of
Proposition~\ref{prop:raw-inconsistency}.  We also report thresholded
summaries for Exponentiated Gradient~\cite{agarwal2018reductions}, Prejudice
Remover~\cite{kamishima2012fairness}, and Adversarial
Debiasing~\cite{zhang2018mitigating}. Complete per-dataset details are
deferred to Appendix~\ref{app:threshold-baselines}, because these
methods do not all share the matched neural in-training backbone used
in the headline Pareto plots.

\begin{figure}[t]
\centering
\begin{subfigure}{0.32\linewidth}
    \centering
    \includegraphics[
      width=\linewidth,
      trim=2.5mm 3mm 1mm 2mm,
      clip
    ]{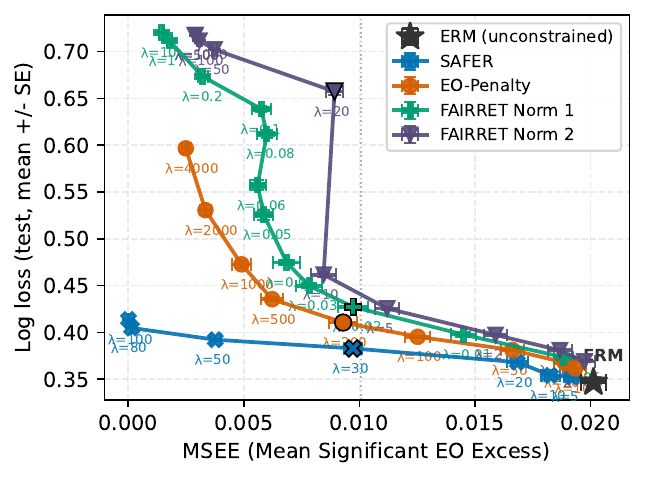}
    \caption{Adult Income}
    \label{fig:pareto-eo-adult}
\end{subfigure}
\hfill
\begin{subfigure}{0.32\linewidth}
    \centering
    \includegraphics[
      width=\linewidth,
      trim=2.5mm 3mm 1mm 2mm,
      clip
    ]{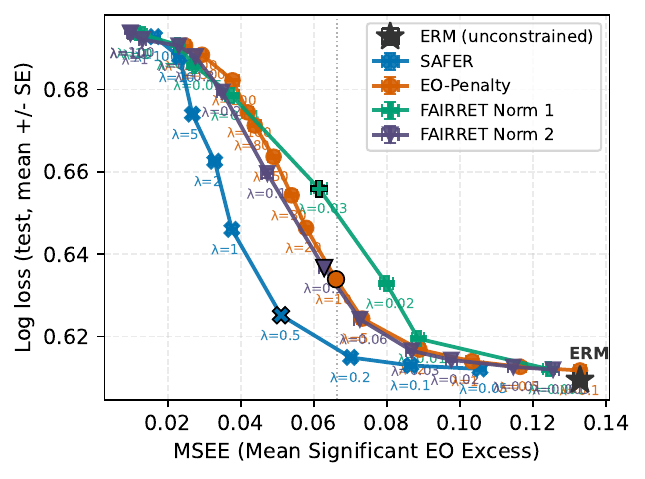}
    \caption{COMPAS}
    \label{fig:pareto-eo-compas}
\end{subfigure}
\hfill
\begin{subfigure}{0.32\linewidth}
    \centering
    \includegraphics[
      width=\linewidth,
      trim=2.5mm 3mm 1mm 2mm,
      clip
    ]{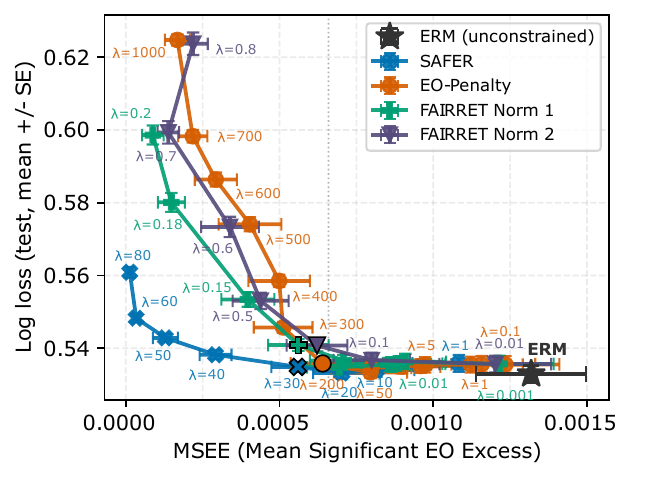}
    \caption{German Credit}
    \label{fig:pareto-eo-german}
\end{subfigure}

\caption{Held-out log-loss versus MSEE Pareto frontiers under Equal
Opportunity.  The vertical axis is the mean test log-loss over
runs across splits and seeds and the horizontal axis is MSEE on hard test
predictions; lower is better on both.  Markers are the values of the
fairness weight $\lambda$ along each method's sweep, annotated and
connected in increasing order, with error bars of $\pm1$ standard error
in both coordinates; the star is ERM ($\lambda=0$).  The methods are
SAFER, EO-Penalty, FAIRRET Norm 1 and FAIRRET Norm 2. Adult Income and German Credit use the default variants
in Table~\ref{tab:datasets}; COMPAS uses the balanced 50/50 variant. The
dashed vertical line marks \(0.5\times\) the ERM MSEE, which is the
eligibility threshold used in Table~\ref{tab:threshold-baseline-summary};
outlined markers indicate the selected qualifying points reported in that
table.}
\label{fig:pareto-eo}
\end{figure}

Table~\ref{tab:threshold-baseline-summary} summarizes the thresholded
fairness--utility comparison across all five datasets and all baselines.
For each dataset and method, the entry is the best mean test log loss among
the swept hyperparameters whose test MSEE is at least 50\% below the ERM MSEE
on the same dataset.  A \(*\) indicates that no swept hyperparameter achieved
that fairness threshold, so the method did not produce a qualifying tradeoff
under this criterion.  Appendix~\ref{app:threshold-baselines} reports the
complete per-dataset tables, including the lowest-MSEE point found for each
ineligible \(*\) case.

\begin{table}[h]
\centering
\small
\vspace{-5pt}
\caption{Thresholded fairness--utility summary across datasets and baselines. Each entry is the best mean test log loss among swept hyperparameters with MSEE at least 50\% below ERM MSEE; \(*\) means no qualifying hyperparameter was found.}
\label{tab:threshold-baseline-summary}
\resizebox{0.95\linewidth}{!}{
\begin{tabular}{llllll}
\toprule
 & Adult Income & COMPAS & German Credit & Taiwan & ACS \\
\midrule
SAFER & \(\mathbf{0.3829} \pm 3\mathrm{e}{-4}\) & \(\mathbf{0.6252} \pm 7\mathrm{e}{-4}\) & \(\mathbf{0.5349} \pm 1\mathrm{e}{-3}\) & \(\mathbf{0.4479} \pm 4\mathrm{e}{-4}\) & \(\mathbf{0.5198} \pm 3\mathrm{e}{-4}\) \\
EO-Penalty & \(0.4106 \pm 9\mathrm{e}{-4}\) & \(0.6339 \pm 5\mathrm{e}{-4}\) & \(0.5358 \pm 1\mathrm{e}{-3}\) & \(0.4527 \pm 6\mathrm{e}{-4}\) & \(0.5223 \pm 4\mathrm{e}{-4}\) \\
FAIRRET Norm 1 & \(0.4272 \pm 1\mathrm{e}{-3}\) & \(0.6559 \pm 9\mathrm{e}{-4}\) & \(0.5409 \pm 2\mathrm{e}{-3}\) & \(0.4668 \pm 1\mathrm{e}{-3}\) & \(0.5240 \pm 4\mathrm{e}{-4}\) \\
FAIRRET Norm 2 & \(0.4618 \pm 3\mathrm{e}{-3}\) & \(0.6366 \pm 7\mathrm{e}{-4}\) & \(0.5408 \pm 2\mathrm{e}{-3}\) & \(0.4624 \pm 1\mathrm{e}{-3}\) & \(0.5314 \pm 4\mathrm{e}{-4}\) \\
AdvDebiasing & * & * & * & * & * \\
ExponGrad & * & * & * & \(0.6256 \pm 3.5\mathrm{e}{-2}\) & \(0.6535 \pm 1.3\mathrm{e}{-2}\) \\
PrejudiceRemover & * & * & * & \(0.6216 \pm 1.1\mathrm{e}{-2}\) & * \\
\bottomrule
\end{tabular}
}
\end{table}

Figure~\ref{fig:pareto-eo} shows the EO Pareto frontiers for Adult,
COMPAS, and German.
Across panels, SAFER preserves near-ERM log-loss while
crossing the threshold and reaches lower MSEE than EO-Penalty or
FAIRRET without comparable log-loss collapse.
Both observations are consistent with the alignment argument of
Theorem~\ref{thm:thresh-consistency}: optimizing a threshold-smoothed
surrogate that approximates the hard-decision metric with an error
linear in $\tau$ is more
effective at reducing that metric than optimizing a raw-soft-count
surrogate, especially in regions where the soft-count and
hard-decision quantities can disagree by an order-one amount
(Proposition~\ref{prop:raw-inconsistency}).
SAFER's frontier matches or improves on that of a baseline that
optimizes the Wald statistic directly (Appendix~\ref{app:wald-stat}).
The per-dataset threshold summaries, including the minimum-MSEE points
for methods marked ineligible in Table~\ref{tab:threshold-baseline-summary},
are reported in Appendix~\ref{app:threshold-baselines}.

The corresponding SP Pareto plots are in Appendix~\ref{app:sp}.  Under
SP the effective per-subgroup sample sizes \(n_S\) are substantially
larger than EO's \(m_S\), so more subgroups are certified already at ERM
and the gate contributes less.  SAFER nevertheless reaches lower log-loss
at matched MSPE than the raw-probability penalty on all three datasets,
and the difference is largest on Adult.  We attribute this to the
threshold-smoothing effect of
Theorem~\ref{thm:thresh-consistency}: with the gate close to open, the
hinge approaches a smoothed-disparity penalty, and SAFER's remaining
advantage comes from targeting hard-decision disparities.

\subsection{Ablation study}
\label{sec:exp-ablation}

Section~\ref{sec:safer} predicted three mechanisms: threshold
smoothing (Theorems~\ref{thm:thresh-consistency},
\ref{thm:gradient-localization}), the gate threshold
(Proposition~\ref{cor:safety}), and the dense tail
(Eq.~\eqref{eq:hinge-suppression}).  Each picks out one stress variant
in which it is predicted to be empirically necessary.  We test all
three with single-knob ablations against the SAFER backbone
(NoThreshSmooth, NoSelectivity, NoDenseTail);
details are in
Appendix~\ref{app:ablation-details}.

\begin{figure}[t]
\centering
\includegraphics[
  width=0.325\linewidth,
  trim=2.5mm 3mm 1mm 2mm,
  clip
]{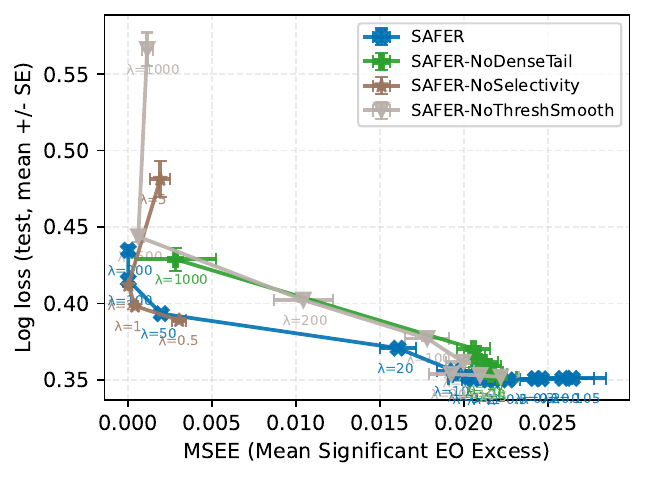}
\includegraphics[
  width=0.325\linewidth,
  trim=2.5mm 3mm 1mm 2mm,
  clip
]{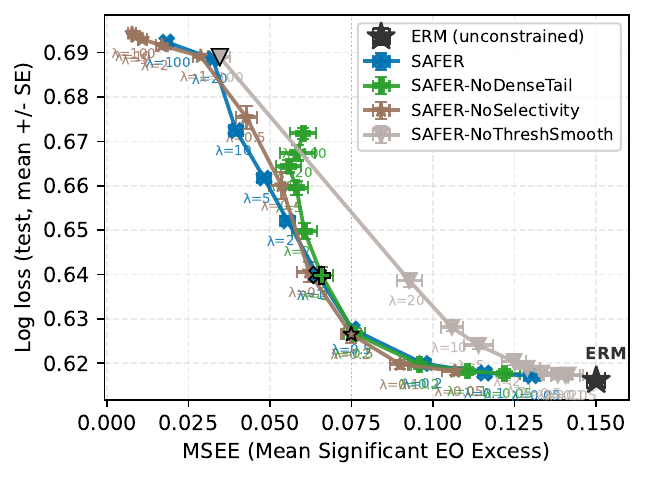}
\includegraphics[
  width=0.325\linewidth,
  trim=2.5mm 3mm 1mm 2mm,
  clip
]{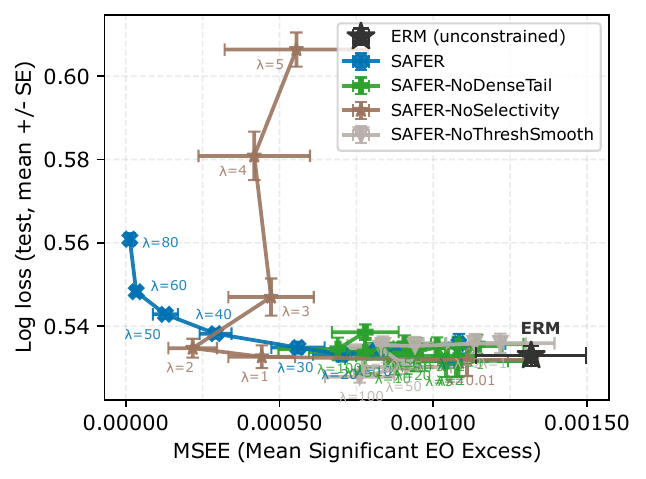}
\caption{Single-knob ablations of SAFER on the default Adult Income,
COMPAS, and German Credit variants from Table~\ref{tab:datasets}.
The panels show Adult Income (left), COMPAS (center) and German
Credit (right).  Axes, markers and error bars are as in
Figure~\ref{fig:pareto-eo}.  The arms are SAFER and its three
single-knob removals: NoDenseTail ($w_{\mathrm{tail}}=0$),
NoSelectivity (gate threshold set to zero) and NoThreshSmooth
(raw probabilities instead of threshold-smoothed counts).}
\label{fig:ablation-compas-german}
\vspace{-5pt}
\end{figure}

\paragraph{Threshold smoothing.}
On Adult and COMPAS, NoThreshSmooth is
Pareto-dominated across the mid-MSEE range; at matched
\(MSEE\!\approx\!10^{-3}\) on Adult, removing threshold smoothing
costs roughly \(0.14\) nats of log-loss.  This is the empirical
counterpart of Theorems~\ref{thm:thresh-consistency}
and~\ref{thm:gradient-localization}: raw probabilities target a
soft-count quantity that is biased relative to the hard-decision
metric and whose gradient is not concentrated where decisions can
flip.

\paragraph{Dense-tail term.}
On COMPAS, NoDenseTail plateaus at
\(MSEE\!\approx\!0.06\) from \(\lambda\!\ge\!5\) while SAFER reaches
\(MSEE\!\approx\!0.018\) at \(\lambda\!=\!100\) (about ten standard
errors apart).  On Adult, NoDenseTail similarly fails to
reach the low-MSEE region of SAFER's frontier: it remains near ERM
MSEE (\(\approx 0.02\)) for almost the entire \(\lambda\) sweep and
moves toward smaller MSEE only at the largest tested \(\lambda\),
where it is Pareto-dominated by SAFER
(Figure~\ref{fig:ablation-compas-german}, left).  The mechanism is
Eq.~\eqref{eq:hinge-suppression}: as the optimizer drives disparities
toward the gate threshold, \(\sigma(u_S/\tau_s)\) drops
below \(\tfrac12\) and the hinge becomes a vanishing fraction of the
gradient; the tail's gradient carries no \(\sigma(u_S/\tau_s)\)
factor and lets optimization continue.  On German Credit,
NoDenseTail tracks SAFER, because the significance gate already
keeps the hinge inactive.

\paragraph{Gate threshold.}
On German Credit, almost no EO violation is statistically certified at
\(\alpha=0.05\) in our sweeps; SAFER preserves ERM-level log-loss
(\(\approx 0.53\)) up to \(\lambda\!=\!20\), while
NoSelectivity climbs to log-loss \(\approx 0.69\) by
\(\lambda\!=\!20\).  This is the failure mode that
Proposition~\ref{cor:safety} predicts when the optimizer can act on
threshold-smoothed but uncertified signal.

\spara{Cross-Regime Summary.}
\label{sec:exp-summary}
Each removal produces the failure that Section~\ref{sec:safer}
predicts.  Without threshold smoothing, a Pareto gap opens on Adult and
COMPAS, as Theorems~\ref{thm:thresh-consistency}
and~\ref{thm:gradient-localization} predict.  Without the dense tail, the
frontier stops short of low MSEE on the same two datasets, because of the
hinge suppression in Eq.~\eqref{eq:hinge-suppression}.  Without the gate
threshold, the log-loss rises on German Credit, as
Proposition~\ref{cor:safety} predicts.  NoSelectivity has no
effect on COMPAS, and on Adult it tracks SAFER up to \(\lambda=1\) and
loses utility at larger \(\lambda\)
(Appendix~\ref{app:selective-gating-real}).  The ablation map is in
Appendix~\ref{app:cross-regime}.

\section{Conclusion}
\label{sec:conclusion}

Intersectional fairness auditing encounters a resolution limit: as subgroup definitions grow finer, the statistical uncertainty of each disparity estimate comes to exceed the disparity itself.
We showed that ignoring this limit, as existing in-processing methods do, leads to fairness overfitting, where the optimizer spends model capacity fitting sampling noise rather than genuine disparities.

To address this, we introduced the \emph{significant-excess} metric family, which continuously measures only the portion of each subgroup's disparity that exceeds what sampling variation can explain at a given confidence level, and \emph{SAFER}, a differentiable regularizer that gates fairness penalties on a soft-count likelihood-ratio evidence statistic.
SAFER satisfies a provable safety property: when every subgroup disparity lies well below the gate threshold, the ERM optimum is nearly stationary for SAFER. At that optimum, the gated fairness contribution is exponentially small, and the contribution of the small ungated term is proportional to its weight.

Experiments across several benchmarks and purpose-built stress variants support a regime-conditional claim: SAFER achieves a better Pareto frontier than the baselines where certified disparities exist and preserves the utility of the unconstrained model where they do not, and removing any component of SAFER produces the failure the theory predicts on at least one stress variant.

\spara{Limitations and future work.}
A zero significant excess means that no excess disparity is
certifiable at the chosen confidence level and sample size; a real
disparity in a small subgroup can remain uncertified, and SAFER then
withholds the gradient that would reduce it
(Appendix~\ref{app:low-power}).
The current evaluation is limited to tabular benchmarks with at most a few hundred intersectional subgroups.
Scaling to high-cardinality attribute spaces and to settings where subgroup membership is uncertain or partially observed remains open.
A natural extension is to include multiplicity corrections in the training-time significance test, so that SAFER's gate also controls the family-wise error rate.

\begin{ack}
This work is conducted as part of the Horizon Europe project PRE-ACT (Prediction of Radiotherapy side effects using explainable AI for patient communication and treatment modification). It is supported by the European Commission through the Horizon Europe Program (Grant Agreement number 101057746), by the Swiss State Secretariat for Education, Research and Innovation (SERI) under contract number 22 00058, and by the UK government (Innovate UK) under application number 10061955.
The authors are employed by Intesa Sanpaolo. The authors declare no other competing interests.
\end{ack}

\bibliographystyle{plain}
\bibliography{references_b}

\appendix
\setcounter{figure}{0}
\setcounter{table}{0}
\setcounter{equation}{0}
\renewcommand{\thefigure}{A.\arabic{figure}}
\renewcommand{\thetable}{A.\arabic{table}}
\renewcommand{\theequation}{A.\arabic{equation}}
\renewcommand{\theHfigure}{A.\arabic{figure}}
\renewcommand{\theHtable}{A.\arabic{table}}
\renewcommand{\theHequation}{A.\arabic{equation}}

\section{Datasets and stress variants}
\label{app:datasets}

All experiments use the same preprocessing code as the released scripts.
Protected attributes are removed from the feature matrix and kept only for
subgroup construction.  Subgroup masks include all conjunctions up to the
specified intersection level, with groups of size less than two omitted in the
corresponding split.

\spara{Adult Income.}
We use the OpenML version of the UCI Adult Income dataset~\cite{kohavi1996scaling}.  The
label is income \(>\$50\)K.  The protected attributes are race, sex, age
binned into \(\{<25,25\text{--}45,45\text{--}65,>65\}\), and native country
binarized as US/non-US.  Non-protected numeric and categorical features are
retained, with categorical features one-hot encoded.  The stress variant in
the main paper uses intersection level~4 and a 10/90 train/test split.  After
dropping rows with missing values, the processed data contain 45,222 examples,
55 non-protected features, and a positive-label rate of 24.8\%.  Over 10
stratified splits used for a dataset audit, this dataset variant has 4,522
training examples, \(227.7\pm4.0\) training-set subgroup masks and
\(256.2\pm2.6\) test-set subgroup masks.  This is a budget of only
\(19.9\pm0.4\) training examples per audited subgroup, and only
\(4.9\pm0.1\) positive-label examples per subgroup for EO.  For the ERM model
fit, the test-set EO audit has
\(220.7\pm2.6\) EO-defined subgroups and certifies \(61.4\pm8.8\)
significant EO violations at \(\alpha=0.05\) after BH correction.

\spara{COMPAS.}
We use the OpenML COMPAS recidivism dataset~\citep{angwin2022machine}.  The label is
two-year recidivism.  The protected attributes are race
(African-American/Caucasian in the OpenML version), sex, and age category
(\(<25\), \(25\text{--}45\), \(>45\)).  The processed data contain 4,966
examples, 5 non-protected features, and a positive-label rate of 50.0\%.
The headline Pareto comparison uses intersection level~3 and a 50/50
train/test split.  Over ten stratified splits this gives 2,483 training
examples and exactly 35 audited subgroup masks in both train and test, a
budget of \(70.9\) training examples and \(35.5\) positive-label examples per
audited subgroup.  We run an ERM test-set EO audit with 10 splits that certifies \(27.3\pm2.0\) significant EO violations at \(\alpha=0.05\)
after BH correction.  The mechanism ablation uses the more severe 20/80 split:
993 training examples, the same 35 subgroup masks, and only \(28.4\) training
examples and \(14.2\) positives per audited subgroup.  Under this stress split,
ERM certifies \(29.3\pm2.2\) significant EO violations, making it the
high-regularization stress test for the dense-tail component.

\spara{German Credit.}
We use the OpenML version of the Statlog German Credit dataset~\cite{statlog_german_credit_data_144}.  The label is good credit.
The protected attributes are sex derived from \texttt{personal\_status}, age
binned into \(\{<30,30\text{--}50,>50\}\), and foreign-worker status.  The
processed data contain 1,000 examples, 53 non-protected features, and a
positive-label rate of 70.0\%.  The stress variant uses intersection level~3
and a 50/50 train/test split.  Over ten stratified splits this gives 500
training examples, \(32.1\pm1.4\) training-set subgroup masks, and
\(31.1\pm0.3\) test-set subgroup masks, a budget of \(15.6\pm0.7\) training
examples and \(10.9\pm0.5\) positive-label examples per audited subgroup.
For the ERM model fit on these same splits, the test-set EO
audit has \(31.1\pm0.3\) EO-defined subgroups and certifies only
\(0.3\pm0.9\) significant EO violations at \(\alpha=0.05\) after BH
correction, with MSEE \(0.0007\pm0.0018\).  This is therefore a safety-regime
test rather than a fairness-improvement test.

\spara{ACS Coverage.}
We use the ACS Public Coverage task from the Folktables benchmark~\cite{ding2021retiring}, which
predicts public health insurance coverage from American Community Survey
records.  The protected attributes are race, sex, and age. The reported variant uses
intersection level~3 and a 50/50 train/test split. The data contains 50,000 examples, so each split has 25,000 training examples.
Across the dataset audit there are exactly 71 training-set and test-set
subgroup masks, all EO-defined.  This corresponds to a budget of 352.1
training examples and 103.5 positive-label examples per audited subgroup
(\(\approx 29.4\%\) positives).  Because the subgroup budget is much larger
than in Adult, COMPAS, or German, this variant serves as a higher-sample ACS
coverage benchmark rather than a sparse-cell stress test.

\spara{Taiwan Credit.}
We use the UCI Default of Credit Card Clients dataset
\citep{yeh2009comparisons}, which predicts credit card default.  The protected
attributes are sex, education, marriage, and age.  The headline variant uses
intersection level~4 and a 50/50 train/test split.  The processed data contain
30,000 examples, so each split has 15,000 training examples.  Over ten
stratified splits this gives \(285.4\pm2.8\) training-set subgroup masks and
\(285.9\pm2.3\) test-set subgroup masks.  This corresponds to a budget of
\(52.6\pm0.5\) training examples and \(11.6\pm0.1\) positive-label examples
per audited subgroup; \(245.8\) subgroups are EO-defined on average.  This
variant therefore provides a high-intersectionality credit-risk benchmark with
a larger total training set than Adult but still a relatively small
positive-label budget per audited subgroup.

\section{Training details and hyperparameters}
\label{app:training}

\spara{Model and optimizer.}
The headline neural methods use the same MLP classifier: hidden layers
\((64,32)\), ReLU activations, dropout \(0.2\), Adam optimizer
(\(\mathrm{lr}=10^{-3}\), weight decay \(5\times10^{-4}\)), batch size 256,
gradient clipping at norm 1, and at most 200 epochs.  Prediction loss is binary
cross-entropy.  Fairness statistics are computed from full-training-set soft
counts each epoch rather than mini-batches, because subgroup-level LRT and
significant-excess quantities are unstable on mini-batch counts.

\spara{Splits.}
The main Adult L4 10/90 and COMPAS L3 50/50 Pareto comparisons use \(100\)
stratified shuffle splits and \(3\) random seeds per split.  The mechanism
ablations use 10 splits \(\times\) 3 seeds on Adult L4 10/90,
100 splits \(\times\) 3 seeds on COMPAS L3 20/80, and
30 splits \(\times\) 3 seeds on German L3 50/50.
Within the training portion of every split, \(10\%\) is held out for model
selection.  The held-out test partition is never used for hyperparameter or
checkpoint selection.

\spara{Checkpointing.}
SAFER and its single-knob ablations use the joint BCE-guarded training
configuration.  The selector first restricts attention to checkpoints whose
validation BCE is within a relative tolerance of the best validation BCE, and
then selects the checkpoint with the smallest guarded score, using the
full-training-set fairness loss for the fairness term.  This prevents high
\(\lambda\) runs from selecting collapsed models while still allowing fairness
improvements when their validation-BCE cost is small.  ERM selects by
validation BCE alone.  The headline EO-Penalty and FAIRRET runs use their
standard joint-training selectors unless otherwise stated.

\spara{SAFER hyperparameters.}
The main SAFER configuration is the unweighted subgroup average
(\texttt{use\_size\_weight=false}) with
\(\alpha=0.05\), threshold smoothing temperature
\(\tau=0.05\), softplus temperature \(\tau_s=0.01\),
absolute-value smoothing \(\varepsilon=10^{-3}\), detached LRT-based
\(\widehat{SE}_{\mathrm{eff}}\), and LRT dense-tail weight
\(w_{\mathrm{tail}}=0.003\).  The NoThreshSmooth ablation sets
\(\tau\) to null; NoSelectivity sets the hinge threshold to
zero; NoDenseTail sets \(w_{\mathrm{tail}}=0\).

\begin{table}[t]
\centering
\small
\caption{Regularization grids used in the Results section.  Main Pareto
comparisons use 100 splits \(\times\) 3 seeds; Adult ablations use
10 splits \(\times\) 3 seeds.}
\label{tab:app-grids}
\resizebox{\linewidth}{!}{%
\begin{tabular}{lll}
\toprule
Experiment & Method & Grid \\
\midrule
Adult L4 10/90 main & SAFER
& \(0.05,0.1,0.2,0.3,0.5,1,2,5,10,20,30,50,80,100\) \\
Adult L4 10/90 main & EO-Penalty
& \(10,20,50,100,200,500,1000,2000,4000\) \\
Adult L4 10/90 main & FAIRRET Norm 2
& \(0.01,0.1,0.5,1,2,5,10,20,50,100,500,1000\) \\
Adult L4 10/90 ablation & SAFER / NoThresh / NoTail
& \(5,10,20,50,100,200,500,1000\) \\
Adult L4 10/90 ablation & NoSelectivity
& \(0.5,1,2,5,10,20,50,100\) \\
COMPAS L3 50/50 main & SAFER
& \(0.05,0.1,0.2,0.5,1,2,5,10,100\) \\
COMPAS L3 50/50 main & EO-Penalty
& \(0.1,0.5,1,2,5,10,20,30,50,80,100,200,500,1000\) \\
COMPAS L3 50/50 main & FAIRRET Norm 1
& \(0.001,0.01,0.02,0.03,0.04,0.05,0.1,1,10\) \\
COMPAS L3 50/50 main & FAIRRET Norm 2
& \(0.005,0.01,0.02,0.03,0.06,0.1,0.15,0.2,0.3,0.5,1,10,100\) \\
COMPAS L3 20/80 ablation & SAFER / NoThresh / NoTail
& \(0.05,0.1,0.2,0.5,1,2,5,10,20,100\) \\
COMPAS L3 20/80 ablation & NoSelectivity
& \(0.05,0.1,0.2,0.3,0.4,0.5,1,2,5,10,100\) \\
German L3 50/50 ablation & SAFER / NoThresh / NoTail
& \(1,2,5,10,20,50,100\) \\
German L3 50/50 ablation & NoSelectivity
& \(0.01,1,2,3,4,5,10,20,50,100\) \\
\bottomrule
\end{tabular}
}
\end{table}

\spara{Baselines.}
ERM is the unconstrained MLP.  EO-Penalty is BCE plus an unweighted
squared EO-disparity penalty.  FAIRRET Norm 1 and FAIRRET Norm 2 use the
ratio-normalized differentiable fairness regularizers of FAIRRET~\cite{buyl2024fairret}, with the EO variant conditioning on positive labels.
We also swept ExponGrad~\citep{agarwal2018reductions} with TPR-parity
constraints, Prejudice Remover, and adversarial debiasing where compatible.
These additional baselines are included in the thresholded summary, but they
do not all share the same neural in-training backbone as the headline Pareto
comparisons.

\spara{Uncertainty reporting.}
Figures report means and \(\pm1\) standard error over all combinations of split and seed.
Because three seeds share the same data split, these observations are not
fully independent.  We therefore treat the plotted SE as descriptive and
verified that the method ordering is unchanged when seeds are averaged within
split before computing SE.

\section{Computational resources}
\label{app:compute}

All reported neural sweeps were run on CPU workers.  The machine used for the
final local runs was an Apple MacBook Pro with an Apple M1 Max processor
(10 CPU cores: 8 performance and 2 efficiency cores) and 32~GB of unified
memory, running macOS/Darwin on arm64.  GPU execution was disabled in the
experiment scripts.  Independent \(\lambda\) jobs
were parallelized across up to 8 CPU worker processes.  Each worker trains
one combination of method and \(\lambda\) over its assigned sequence of splits and seeds.
Peak memory was dominated by the Python/PyTorch worker processes and the
dataset matrices; on this 32~GB machine the reported CPU sweeps ran without
memory pressure when running all eight processes.

Table~\ref{tab:app-compute-time} gives wall-clock estimates for reproducing
the full neural sweeps.  These estimates exclude plotting and
table aggregation, which take seconds to minutes, and should be interpreted as
order-of-magnitude wall-clock requirements for this CPU workstation rather
than hardware-independent constants.

\begin{table}[t]
\centering
\small
\caption{Estimated CPU wall-clock time for the full sweeps on the
10-core, 32~GB Apple M1 Max machine with eight joblib workers. SAFER-ablation variants
(\textsc{NoThreshSmooth}, \textsc{NoSelectivity}, and \textsc{NoDenseTail})
have the same resource profile as SAFER up to small constant-factor changes.}
\label{tab:app-compute-time}
\resizebox{\linewidth}{!}{%
\begin{tabular}{llccc}
\toprule
Experiment & Method & \(\lambda\) jobs & Split \(\times\) seed runs per job & Estimated wall time \\
\midrule
Adult L4 10/90 main & SAFER & 14 & \(100\times3\) & 3--6 h \\
Adult L4 10/90 main & EO-Penalty & 9 & \(100\times3\) & 1--1.5 h \\
Adult L4 10/90 main & FAIRRET Norm 1 / Norm 2 & 12--13 & \(100\times3\) & 20--45 min \\
COMPAS L3 50/50 main & SAFER & 9 & \(100\times3\) & 20--40 min \\
COMPAS L3 50/50 main & EO-Penalty & 14 & \(100\times3\) & 15--30 min \\
COMPAS L3 50/50 main & FAIRRET Norm 1 / Norm 2 & 9--13 & \(100\times3\) & 5--10 min \\
German L3 50/50 main & SAFER & 7 & \(100\times3\) & 20--30 min \\
German L3 50/50 main & EO-Penalty & 14 & \(100\times3\) & 10--20 min \\
German L3 50/50 main & FAIRRET Norm 1 / Norm 2 & 9--13 & \(100\times3\) & under 10 min \\
Adult L4 10/90 ablation & SAFER-family arm & 8 & \(30\times3\) & 35--55 min per arm \\
COMPAS L3 20/80 ablation & SAFER-family arm & 10--11 & \(100\times3\) & 20--40 min per arm \\
German L3 50/50 ablation & SAFER-family arm & 7--10 & \(30\times3\) & 5--15 min per arm \\
\bottomrule
\end{tabular}
}
\end{table}

\section{Per-dataset thresholded baseline summaries}
\label{app:threshold-baselines}

The following tables expand Table~\ref{tab:threshold-baseline-summary} with
per-dataset details.  For eligible rows, we report the best mean test log loss
among swept hyperparameters whose MSEE is at least 50\% below the ERM MSEE on
that dataset.  For ineligible rows, corresponding to \(*\) entries in the main
summary table, we report the minimum-MSEE point found by the sweep.

\begin{table}[h!]
\centering
\small
\caption{Adult Income: Best mean test log loss with MSEE at least 50\% below ERM
MSEE. (ERM MSEE=0.0201, threshold=0.0101; ineligible rows show the minimum-MSEE point).}
\label{tab:threshold-adult}
\resizebox{0.95\linewidth}{!}{
\begin{tabular}{llllll}
\toprule
Method & Parameter & Eligible & Log loss & AUC & MSEE \\
\midrule
SAFER & \(\lambda=30\) & yes & \(\mathbf{0.3829}\pm0.0003\) & \(\mathbf{0.8818}\pm0.0003\) & \(0.0097\pm0.0004\) \\
EO-Penalty & \(\lambda=200\) & yes & \(0.4106\pm0.0009\) & \(0.8624\pm0.0006\) & \(0.0093\pm0.0006\) \\
FAIRRET Norm 1 & \(\lambda=0.02\) & yes & \(0.4272\pm0.0012\) & \(0.8561\pm0.0007\) & \(0.0097\pm0.0006\) \\
FAIRRET Norm 2 & \(\lambda=10\) & yes & \(0.4618\pm0.0030\) & \(0.8456\pm0.0022\) & \(0.0085\pm0.0005\) \\
AdvDebiasing & \(\alpha_{\mathrm{adv}}=0.5\) & no & \(0.4770\pm0.0066\) & \(0.8208\pm0.0034\) & \(0.0157\pm0.0009\) \\
ExponGrad & \(\epsilon_{\mathrm{EG}}=0.05\) & no & \(0.5936\pm0.0238\) & \(0.7373\pm0.0066\) & \(0.0335\pm0.0018\) \\
PrejudiceRemover & \(\lambda=5\) & no & \(1.1022\pm0.1035\) & \(0.4367\pm0.0835\) & \(0.0165\pm0.0038\) \\
\bottomrule
\end{tabular}}
\end{table}

\begin{table}[h!]
\centering
\small
\caption{COMPAS: Best mean test log loss with MSEE at least 50\% below ERM (ERM MSEE=0.1292, threshold=0.0646; ineligible rows show the minimum-MSEE point).} 
\label{tab:threshold-compas}
\resizebox{0.95\linewidth}{!}{
\begin{tabular}{llllll}
\toprule
Method & Parameter & Eligible & Log loss & AUC & MSEE \\
\midrule
SAFER & \(\lambda=0.5\) & yes & \(\mathbf{0.6252} \pm 0.0007\) & \(0.7148 \pm 0.0009\) & \(0.0510 \pm 0.0013\) \\
EO-Penalty & \(\lambda=10\) & yes & \(0.6339 \pm 0.0005\) & \(\mathbf{0.7171} \pm 0.0009\) & \(0.0661 \pm 0.0015\) \\
FAIRRET Norm 1 & \(\lambda=0.03\) & yes & \(0.6559 \pm 0.0009\) & \(0.6975 \pm 0.0018\) & \(0.0614 \pm 0.0021\) \\
FAIRRET Norm 2 & \(\lambda=0.1\) & yes & \(0.6366 \pm 0.0007\) & \(0.7117 \pm 0.0010\) & \(0.0628 \pm 0.0015\) \\
AdvDebiasing & \(\alpha_{\mathrm{adv}}=0.5\) & no & \(0.7381 \pm 0.0449\) & \(0.6522 \pm 0.0170\) & \(0.1029 \pm 0.0095\) \\
ExponGrad & \(\epsilon_{\mathrm{EG}}=0.05\) & no & \(0.7211 \pm 0.0061\) & \(0.5291 \pm 0.0096\) & \(0.1259 \pm 0.0045\) \\
PrejudiceRemover & \(\lambda=0.1\) & no & \(0.6595 \pm 0.0037\) & \(0.7066 \pm 0.0023\) & \(0.0939 \pm 0.0019\) \\
\bottomrule
\end{tabular}
}
\end{table}

\begin{table}[h!]
\centering
\small
\caption{German Credit: Best mean test log loss with MSEE at least 50\% below ERM (ERM MSEE=0.0013, threshold=0.0007; ineligible rows show the minimum-MSEE point).}
\label{tab:threshold-german}
\resizebox{0.95\linewidth}{!}{
\begin{tabular}{llllll}
\toprule
Method & Parameter & Eligible & Log loss & AUC & MSEE \\
\midrule
SAFER & \(\lambda=30\) & yes & \(\mathbf{0.5349} \pm 0.0014\) & \(\mathbf{0.7659} \pm 0.0017\) & \(0.0006 \pm 0.0001\) \\
EO-Penalty & \(\lambda=200\) & yes & \(0.5358 \pm 0.0013\) & \(0.7554 \pm 0.0017\) & \(0.0006 \pm 0.0001\) \\
FAIRRET Norm 1 & \(\lambda=0.12\) & yes & \(0.5409 \pm 0.0016\) & \(0.7536 \pm 0.0019\) & \(0.0006 \pm 0.0001\) \\
FAIRRET Norm 2 & \(\lambda=0.4\) & yes & \(0.5408 \pm 0.0016\) & \(0.7544 \pm 0.0018\) & \(0.0006 \pm 0.0001\) \\
AdvDebiasing & \(\alpha_{\mathrm{adv}}=0.1\) & no & \(1.3272 \pm 0.0141\) & \(0.7247 \pm 0.0024\) & \(0.0027 \pm 0.0003\) \\
ExponGrad & \(\epsilon_{\mathrm{EG}}=0.1\) & no & \(0.5664 \pm 0.0033\) & \(0.7371 \pm 0.0027\) & \(0.0018 \pm 0.0002\) \\
PrejudiceRemover & \(\lambda=5\) & no & \(0.8915 \pm 0.0110\) & \(0.6380 \pm 0.0098\) & \(0.0104 \pm 0.0009\) \\
\bottomrule
\end{tabular}
}
\end{table}

\begin{table}[h!]
\centering
\small
\caption{Taiwan: Best mean test log loss with MSEE at least 50\% below ERM (ERM MSEE=0.0060, threshold=0.0030; ineligible rows show the minimum-MSEE point).}
\label{tab:threshold-taiwan}
\resizebox{0.95\linewidth}{!}{
\begin{tabular}{llllll}
\toprule
Method & Parameter & Eligible & Log loss & AUC & MSEE \\
\midrule
SAFER & \(\lambda=2\) & yes & \(\mathbf{0.4479} \pm 0.0004\) & \(\mathbf{0.7552} \pm 0.0008\) & \(0.0021 \pm 0.0002\) \\
EO-Penalty & \(\lambda=5\) & yes & \(0.4527 \pm 0.0006\) & \(0.7531 \pm 0.0010\) & \(0.0025 \pm 0.0002\) \\
FAIRRET Norm 1 & \(\lambda=0.002\) & yes & \(0.4668 \pm 0.0010\) & \(0.7442 \pm 0.0014\) & \(0.0014 \pm 0.0002\) \\
FAIRRET Norm 2 & \(\lambda=0.05\) & yes & \(0.4624 \pm 0.0012\) & \(0.7455 \pm 0.0015\) & \(0.0021 \pm 0.0002\) \\
AdvDebiasing & \(\alpha_{\mathrm{adv}}=5\) & no & \(0.4569 \pm 0.0009\) & \(0.7546 \pm 0.0013\) & \(0.0032 \pm 0.0002\) \\
ExponGrad & \(\epsilon_{\mathrm{EG}}=0.1\) & yes & \(0.6256 \pm 0.0347\) & \(0.6096 \pm 0.0101\) & \(0.0027 \pm 0.0006\) \\
PrejudiceRemover & \(\lambda=0.01\) & yes & \(0.6216 \pm 0.0107\) & \(0.6093 \pm 0.0103\) & \(0.0017 \pm 0.0002\) \\
\bottomrule
\end{tabular}
}
\end{table}

\begin{table}[h!]
\centering
\small
\caption{ACS: Best mean test log loss with MSEE at least 50\% below ERM (ERM MSEE=0.0904, threshold=0.0452; ineligible rows show the minimum-MSEE point).}
\label{tab:threshold-acs}
\resizebox{0.95\linewidth}{!}{
\begin{tabular}{llllll}
\toprule
Method & Parameter & Eligible & Log loss & AUC & MSEE \\
\midrule
SAFER & \(\lambda=0.3\) & yes & \(\mathbf{0.5198} \pm 0.0003\) & \(\mathbf{0.7472} \pm 0.0006\) & \(0.0236 \pm 0.0009\) \\
EO-Penalty & \(\lambda=10\) & yes & \(0.5223 \pm 0.0004\) & \(0.7427 \pm 0.0007\) & \(0.0383 \pm 0.0009\) \\
fairret-norm1 & \(\lambda=0.005\) & yes & \(0.5240 \pm 0.0004\) & \(0.7401 \pm 0.0006\) & \(0.0357 \pm 0.0011\) \\
fairret-norm2 & \(\lambda=0.05\) & yes & \(0.5314 \pm 0.0004\) & \(0.7320 \pm 0.0007\) & \(0.0218 \pm 0.0009\) \\
AdvDebiasing & \(\alpha_{\mathrm{adv}}=0.5\) & no & \(0.5299 \pm 0.0111\) & \(0.7450 \pm 0.0030\) & \(0.0605 \pm 0.0025\) \\
ExponGrad & \(\epsilon_{\mathrm{EG}}=0.005\) & yes & \(0.6535 \pm 0.0134\) & \(0.5623 \pm 0.0078\) & \(0.0257 \pm 0.0029\) \\
PrejudiceRemover & \(\lambda=10\) & no & \(0.6196 \pm 0.0019\) & \(0.6401 \pm 0.0018\) & \(0.0464 \pm 0.0013\) \\
\bottomrule
\end{tabular}
}
\end{table}

\section{Subgroups below the certification boundary}
\label{app:low-power}
A zero significant excess for a subgroup states only that the
available sample cannot certify an excess at the chosen confidence
level.  The level $\alpha$ sets the trade between reacting to sampling
noise and leaving low-power disparities uncorrected; a larger $\alpha$
protects smaller groups at a utility cost.  The size sweep of
Appendix~\ref{app:synthetic} makes the dependence concrete: the same
planted gap of $-0.15$ is certified in about 9\% of runs at $m_S=5$ and
in about 95\% at $m_S=120$.  We therefore recommend reporting MSEE
together with raw subgroup disparities, confidence intervals, subgroup
counts and a power or minimum-detectable-effect analysis, and treating an
uncertified subgroup as unresolved; possible responses are collecting
more data, pooling groups coarsely or hierarchically, or policy
constraints that do not rely on statistical significance alone.  Where
harm is suspected in a subgroup too small to certify, the appropriate
evidence is of a different kind: the structure of the decision
procedure, the features it uses, the causal path from the protected
attribute to the outcome, and documented practice.  A complete audit
combines statistical outcome testing with this process evidence, and the
certification boundary marks where the hand-over occurs.  Medical-imaging
benchmarks whose small subgroups carry genuine disparities, such as
Harvard-FairVision30k~\cite{luo2023fairvision}, are a natural next test
of SAFER outside the tabular setting.

\section{Additional experiments}
\label{app:additional}

\subsection{Decision-threshold sensitivity}
\label{app:threshold-sweep}
The threshold $t$ does not enter the prediction loss, which trains
probabilistic scores; it defines the deployed classifier and therefore
both its accuracy and its hard-decision fairness audit.  We use $t=0.5$
because, when scores estimate class probabilities and false positives and
false negatives cost the same, it is the Bayes-optimal decision threshold.
Since $t$ enters SAFER's objective, it is fixed before training: by a
deployment requirement such as a cost matrix or a regulatory decision
rule, or, when no policy value exists, on validation data with a
preliminary ERM model, and then held fixed while all methods are trained
and compared.  To test the dependence on this default we retrained SAFER
for every $t\in\{0.1,\dots,0.9\}$ on the three datasets of the main
Pareto comparison, five seeds per threshold, and audited each model at
its own threshold.  Table~\ref{tab:threshold-sweep} reports the test
log-loss, the paired difference
$\Delta\mathrm{MSEE}=\mathrm{MSEE}_{\mathrm{SAFER}}-\mathrm{MSEE}_{\mathrm{ERM}}$
against the ERM audit at the same threshold (negative values are a
fairness improvement), and the accuracy.  Log-loss is computed from
probabilistic scores, so for a fixed model it does not depend on $t$; its
variation across columns measures how $t$ changes the retrained model
through the fairness term.  Log-loss is stable across $t$ on Adult and
German and more sensitive on COMPAS.  The fairness benefit persists across
operationally meaningful thresholds: Adult has non-positive
$\Delta\mathrm{MSEE}$ throughout, COMPAS improves for $t$ between $0.2$
and $0.9$, and the changes on German are negligible because little
certified disparity is available to reduce.  Extreme thresholds produce
nearly constant classifiers with trivially low MSEE (accuracy $0.500$ on
COMPAS for $t\le0.3$, $0.700$ on German for $t\le0.2$), so those columns
are not useful operating points.  This is a sensitivity analysis; $t$ is
never tuned on the test set.

\begin{table}[h]
\centering\small
\caption{SAFER retrained and audited at decision thresholds $t$ from
$0.1$ to $0.9$: test log-loss, paired $\Delta$MSEE against ERM at the same
threshold, and accuracy (mean $\pm$ standard deviation over five seeds).}
\label{tab:threshold-sweep}
\resizebox{\linewidth}{!}{%
\begin{tabular}{llccccccccc}
\toprule
 & $t$ & 0.1 & 0.2 & 0.3 & 0.4 & 0.5 & 0.6 & 0.7 & 0.8 & 0.9 \\
\midrule
Adult & log-loss & $.336\pm.002$ & $.337\pm.002$ & $.337\pm.003$ & $.336\pm.002$ & $.336\pm.002$ & $.336\pm.002$ & $.337\pm.002$ & $.338\pm.003$ & $.337\pm.003$ \\
 & $\Delta$MSEE & $-.008\pm.009$ & $-.013\pm.005$ & $-.001\pm.009$ & $-.007\pm.006$ & $-.009\pm.006$ & $-.008\pm.003$ & $-.014\pm.004$ & $-.010\pm.005$ & $-.008\pm.005$ \\
 & accuracy & $.696\pm.004$ & $.778\pm.003$ & $.822\pm.003$ & $.839\pm.002$ & $.843\pm.002$ & $.837\pm.002$ & $.817\pm.002$ & $.785\pm.004$ & $.762\pm.002$ \\
\midrule
COMPAS & log-loss & $.615\pm.008$ & $.615\pm.007$ & $.623\pm.006$ & $.633\pm.005$ & $.640\pm.008$ & $.643\pm.007$ & $.628\pm.005$ & $.620\pm.005$ & $.615\pm.005$ \\
 & $\Delta$MSEE & $.000\pm.000$ & $-.005\pm.004$ & $-.089\pm.019$ & $-.131\pm.021$ & $-.084\pm.018$ & $-.069\pm.018$ & $-.078\pm.012$ & $-.051\pm.015$ & $-.011\pm.005$ \\
 & accuracy & $.500\pm.000$ & $.500\pm.000$ & $.500\pm.000$ & $.535\pm.040$ & $.631\pm.016$ & $.531\pm.024$ & $.510\pm.006$ & $.505\pm.003$ & $.502\pm.001$ \\
\midrule
German & log-loss & $.531\pm.035$ & $.531\pm.035$ & $.532\pm.035$ & $.532\pm.035$ & $.531\pm.035$ & $.531\pm.035$ & $.531\pm.035$ & $.532\pm.035$ & $.531\pm.035$ \\
 & $\Delta$MSEE & $.0000\pm.0000$ & $.0000\pm.0000$ & $.0000\pm.0000$ & $-.0005\pm.0013$ & $+.0002\pm.0014$ & $-.0005\pm.0007$ & $-.0014\pm.0018$ & $-.0041\pm.0078$ & $-.0006\pm.0027$ \\
 & accuracy & $.700\pm.000$ & $.700\pm.000$ & $.702\pm.004$ & $.722\pm.018$ & $.731\pm.019$ & $.737\pm.014$ & $.673\pm.109$ & $.590\pm.162$ & $.478\pm.103$ \\
\bottomrule
\end{tabular}}
\end{table}

\subsection{A baseline that optimizes the Wald statistic}
\label{app:wald-stat}
Wald-Stat penalizes the standardized Wald statistic directly:
\begin{equation}
\mathcal{L}_{\mathrm{Wald\text{-}Stat}}
=
\mathcal{L}_{\mathrm{pred}}
+
\frac{\lambda}{|\mathfrak S|}
\sum_{S\in\mathfrak S}
\left[
\operatorname{softplus}_{\tau_s}\left(|Z_S|-z_\alpha\right)^2
+
w_{\mathrm{tail}}|Z_S|
\right],
\qquad
Z_S
=
\frac{\widehat\phi^\tau_S}{\widehat{SE}^\phi_S},
\end{equation}
where $\widehat\phi^\tau_S$ is the disparity computed from the same
threshold-smoothed counts as SAFER and $\widehat{SE}^\phi_S$ its
two-proportion plug-in standard error (Section~\ref{sec:safer-lrt}).
Both stay in the computation graph, so gradients flow through the
statistic, including the route that lowers the penalty by changing the
estimated standard error.  Wald-Stat uses the same critical value,
threshold smoothing, temperatures, dense-tail weight, architecture,
optimizer and checkpointing protocol as SAFER; it tests whether
optimizing Wald significance directly yields a stronger fairness--utility
frontier than SAFER's LRT-based construction.  Because the two
objectives live on different scales, we sweep $\lambda$ independently for
each and match operating points by mean test log-loss after examining
both sweeps; Table~\ref{tab:wald-stat} summarizes the matched pairs.
Entries are SAFER/Wald-Stat means $\pm$ standard deviations over
independent runs over splits and seeds, 10 for Adult and COMPAS and 5 for German,
rounded to three decimals; several differences are small relative to
this variability, so we read the frontier rather than individual rows.
For the following table, the ERM reference values in the caption come
from these same runs, a lighter protocol than the 100 splits with three
seeds of the main sweeps, so they differ from the ERM values of the main
tables.

\begin{table}[h]
\centering\small
\caption{SAFER versus Wald-Stat at matched test log-loss.  Each row
pairs a SAFER and a Wald-Stat fairness weight
$(\lambda_{\mathrm S},\lambda_{\mathrm W})$ with similar mean test
log-loss; entries are SAFER/Wald-Stat, mean $\pm$ standard deviation over
runs.  ERM log-loss/MSEE: Adult $0.334/0.041$, COMPAS
$0.613\pm0.009/0.127\pm0.027$, German $0.531\pm0.035/0.006\pm0.005$.}
\label{tab:wald-stat}
\begin{tabular}{llcc}
\toprule
 & $(\lambda_{\mathrm S},\lambda_{\mathrm W})$ & log-loss S/W & MSEE S/W \\
\midrule
Adult & $(1,0.0001)$ & $.337\pm.003/.338\pm.004$ & $.028\pm.013/.040\pm.013$ \\
 & $(5,0.015)$ & $.344\pm.004/.344\pm.005$ & $.028\pm.015/.035\pm.013$ \\
 & $(15,0.5)$ & $.372\pm.005/.372\pm.005$ & $.025\pm.011/.025\pm.014$ \\
 & $(30,2)$ & $.387\pm.003/.391\pm.003$ & $.005\pm.010/.002\pm.007$ \\
 & $(50,3)$ & $.397\pm.003/.397\pm.004$ & $.0001\pm.0002/.0003\pm.0009$ \\
\midrule
COMPAS & $(0.1,0.001)$ & $.615\pm.009/.614\pm.009$ & $.086\pm.026/.087\pm.025$ \\
 & $(0.5,0.005)$ & $.625\pm.010/.621\pm.008$ & $.058\pm.013/.067\pm.017$ \\
 & $(1,0.015)$ & $.647\pm.012/.648\pm.007$ & $.041\pm.015/.052\pm.019$ \\
 & $(3,0.05)$ & $.671\pm.008/.669\pm.006$ & $.036\pm.026/.046\pm.030$ \\
 & $(20,1)$ & $.690\pm.003/.690\pm.003$ & $.040\pm.048/.038\pm.050$ \\
\midrule
German & $(0.25,0.0005)$ & $.531\pm.035/.531\pm.035$ & $.006\pm.005/.006\pm.005$ \\
 & $(1,0.01)$ & $.532\pm.035/.531\pm.035$ & $.007\pm.005/.008\pm.006$ \\
 & $(15,0.02)$ & $.532\pm.032/.532\pm.035$ & $.006\pm.006/.007\pm.006$ \\
 & $(20,0.1)$ & $.533\pm.032/.533\pm.034$ & $.004\pm.004/.005\pm.005$ \\
 & $(40,3)$ & $.540\pm.033/.541\pm.030$ & $.0005\pm.0007/.0015\pm.0021$ \\
\bottomrule
\end{tabular}
\end{table}

Diagnostics on Adult, stratified by the positive-label count $m_S$,
explain the gap.  Across the five matched points, the median of
$|Z_S|/\sqrt{\Lambda_S}$ is $1.00$ for $m_S\ge30$.  For $m_S<10$,
saturation drives the plug-in standard error toward zero and the ratio's
90th percentile is $37$ to $43$, a roughly fortyfold inflation in the most
extreme cases.  Among about $1{,}491$ pairs of subgroup and seed per point,
Wald-Stat flags 41 to 57 disparities that the LRT does not, and never the
reverse.  These saturated groups are 2.7\% of the total but absorb 99\% of
Wald-Stat's penalty and 86\% of its final-epoch gradient over the smallest
subgroups ($m_S<30$), against 43\% for SAFER.  At the near-ERM Adult
point, the residual certified excess of the groups with $m_S>300$ is
$0.0029$ for SAFER and $0.0214$ for Wald-Stat, so SAFER preserves capacity
for the better-supported groups.  SAFER also has lower mean MSEE near ERM
on Adult and at the intermediate COMPAS and German points, with gaps that
are modest relative to the variability across splits and seeds; under strong
regularization both methods converge near zero MSEE at a greater utility
cost.

\subsection{A controlled benchmark for the selective-gating regime}
\label{app:synthetic}
The operating regimes of Section~\ref{sec:safer-safety} suggest a
hypothesis for the gate threshold: when both active and
inactive subgroups are present, the gate gives SAFER its largest
advantage over NoSelectivity, because it withholds penalties from
unsupported disparities while keeping those with support.  The benchmark
below tests it by separating statistical evidence from disparity
magnitude.  It has one large reference cell and 34 probe cells.  The
generator is calibrated so that its population decision rule has a
true-positive rate of $0.85$ on the reference cell and $0.70$ on every
unfair probe, a planted reference gap of $-0.15$; null probes have zero
gap.  Only the positive-label count $m_S$ of the unfair probes varies,
and the reference count is fixed and large.  Group identity is hidden
from the predictor and used only by the fairness loss and the audit,
which keep the $S$-versus-$\bar S$ comparison of the main text.  We
summarize each evidence stratum by the certification statistic
$\sqrt{\Lambda_S}$ computed on the hard predictions of the final ERM
model on the training split, the same statistic as in the size sweep
below (Table~\ref{tab:synthetic-strata}).  The null and strong-evidence
stratum means lie on opposite sides of the gate in every one of the 10
runs, so the benchmark is in the selective-gating regime; cell by cell
the picture is noisier, with at least one null cell certified in 6 runs
and one strong-evidence cell uncertified in 1 run.

\begin{table}[h]
\centering\small
\caption{Evidence strata of the probe cells for the final ERM model:
cells per stratum, planted gap, mean positive-label count $m_S$ in the
training split, mean certification statistic $\sqrt{\Lambda_S}$ on hard
training predictions, and fraction of cells certified
($\sqrt{\Lambda_S}>1.96$); per-run stratum means, reported as mean
$\pm$ standard deviation over 10 runs.}
\label{tab:synthetic-strata}
\begin{tabular}{lccccc}
\toprule
probe stratum & cells & planted gap & $m_S$ & mean $\sqrt{\Lambda_S}$ & certified fraction \\
\midrule
null & 10 & $0$ & $105$ & $.90\pm.16$ & $.07\pm.07$ \\
weak evidence & 12 & $-0.15$ & $6$ & $1.11\pm.25$ & $.11\pm.07$ \\
boundary & 8 & $-0.15$ & $34$ & $1.77\pm.29$ & $.45\pm.13$ \\
strong evidence & 4 & $-0.15$ & $204$ & $4.72\pm.50$ & $.98\pm.08$ \\
\bottomrule
\end{tabular}
\end{table}

\paragraph{Generator and protocol.}
Every cell draws labels $y\sim\mathrm{Bernoulli}(0.5)$.  One signal
feature is distributed as $\mathcal N(\Phi^{-1}(\mathrm{TPR}_{\mathrm{cell}}),1)$
for positives and $\mathcal N(-\Phi^{-1}(0.85),1)$ for negatives, where
$\mathrm{TPR}_{\mathrm{cell}}$ is the planted true-positive rate of the
cell and $\Phi$ is the standard normal distribution function; eight
pure-noise $\mathcal N(0,1)$ features are appended.  The reference cell
has $45{,}000$ examples and planted rate $0.85$, which pins the decision
threshold of a group-blind classifier so that each cell recovers its
planted rate and the complement of every probe has a true-positive rate
close to $0.85$.  A probe cell with design size $m$ has $4m$ examples,
so about $0.9m$ positives fall in the training split; the design sizes
are $10,10,20,20,40,40,120,120,400,400$ for the ten null probes (planted
rate $0.85$), $5,5,6,6,7,7,8,8,5,6,7,8$ for the twelve weak-evidence
probes, $30,30,35,35,40,40,45,45$ for the eight boundary probes and
$150,150,300,300$ for the four strong-evidence probes (planted rate
$0.70$).  Each of the 10 runs draws a fresh dataset of $54{,}832$
examples, a fresh split and a fresh initialization; within a run all
arms share the initialization and the dropout stream, so comparisons
are paired.  Half of the data is held out as test; of the rest, $90\%$
trains the model and $10\%$ is set aside for validation and not used.
The model is an MLP with hidden layers $(64,32)$ and dropout $0.2$,
trained full-batch for $120$ epochs with Adam (learning rate $10^{-3}$,
weight decay $5\times10^{-4}$); the final model is evaluated, without
early stopping or checkpoint selection, so that $\lambda$ traces the
whole fairness--utility frontier.  SAFER uses the values of the main
experiments ($\alpha=0.05$, $\tau=0.05$, $\tau_s=0.01$,
$\varepsilon=10^{-3}$, $w_{\mathrm{tail}}=0.003$); NoSelectivity removes
only the gate and EO-Penalty adds $\lambda$ times the squared EO gap to
the prediction loss.  The $\lambda$ grid is chosen separately for each
arm, from near-ERM to a near-constant predictor.  Every cell is scored
on its training and its test members with hard predictions at $0.5$; a
cell is skipped on a split where it or its complement has fewer than two
positives, which affects only the weak-evidence stratum (10 to 12 of its
12 cells scored per run).  Stratum values are cell averages, reported as
mean $\pm$ standard deviation over the 10 runs.  The generator, the
harness and the per-run results are released with the code.

We compare SAFER with NoSelectivity (NS), which removes only the
gate, and with EO-Penalty (EO), over independent $\lambda$ sweeps.
Table~\ref{tab:synthetic-sweeps} reports the complete sweeps: for every
arm and $\lambda$, the test log-loss and the mean residual certified
excess in each evidence stratum.  Points whose mean test log-loss
reaches $0.5$ approach the log-loss $0.693$ of a constant predictor,
and their near-zero excess comes from degenerate predictions.  From the
remaining part of the frontiers we chose three approximately matched
mean test-log-loss levels, reported in Table~\ref{tab:synthetic-results}
with the fairness weight and test log-loss of each arm and the mean
residual certified excess in each evidence stratum, in percentage
points, as means $\pm$ standard deviations over 10 runs.  The methods
overlap at the near-ERM level A.  At B and C, SAFER has the lowest mean
excess in the boundary and strong-evidence strata; in original units,
SAFER's C values in these two strata are $0.0079$ and $0.0227$, against
$0.0184$ and $0.0465$ for NS and $0.0108$ and $0.0429$ for EO.  SAFER's null-stratum mean at C is higher and the
standard deviations overlap, so the evidence is directional, with no
statistically significant pairwise differences at matched utility.
The full sweeps show the same pattern: EO's strong-stratum excess stays
near $4$ percentage points from $\lambda=50$ to $\lambda=600$ while its
log-loss rises from $0.47$ to $0.63$, and NS becomes unstable above
$\lambda\approx1$, with large run-to-run variation of the log-loss from
$\lambda=1.25$ on.  The strong-evidence gain is an indirect effect of
the whole objective: the gate is already active for those probes, and
withholding unsupported gradients preserves optimization capacity for
well-supported ones.  Table~\ref{tab:synthetic-absphi} reports the raw
gap $|\phi_S|$ on training and on test members for the non-degenerate
points.  On the weak-evidence stratum, relative to ERM (training gap
$0.185$, test gap $0.200$), EO at $\lambda=50$ lowers the training gap by
$0.090$ but the test gap by only $0.037$, whereas SAFER at $\lambda=15$,
at similar log-loss, lowers them by $0.098$ and $0.071$: the
certainty-blind penalty fits training-set noise in the sparse cells.
An independent size sweep (Table~\ref{tab:synthetic-size}) locates the
transition for this planted gap: the certified fraction rises from
$0.09\pm0.09$ at design size $m=5$ to $0.52\pm0.15$ at $m=30$, where
the mean statistic equals $1.96$, and to $0.95\pm0.08$ at $m=120$; the
value $m\approx30$ is specific to this gap and reference size.  On the
real datasets, Adult and COMPAS were built as candidates for this
regime; NoSelectivity tracks SAFER across the sweep on COMPAS and
up to $\lambda=1$ on Adult, where it then loses utility
(Appendix~\ref{app:selective-gating-real}).

\begin{table}[h]
\centering\small
\caption{Matched-utility comparison on the controlled benchmark:
fairness weight $\lambda$, test log-loss, and mean residual certified
excess per evidence stratum in percentage points (mean $\pm$ standard
deviation over 10 runs; lower is better).}
\label{tab:synthetic-results}
\resizebox{\linewidth}{!}{%
\begin{tabular}{llcccccc}
\toprule
set & arm & $\lambda$ & test log-loss & null & weak & boundary & strong \\
\midrule
-- & ERM & $0$ & $.356\pm.004$ & $.89\pm1.03$ & $3.49\pm2.69$ & $2.01\pm.74$ & $7.42\pm1.30$ \\
\midrule
A & SAFER & $2$ & $.362\pm.003$ & $1.09\pm.77$ & $3.35\pm2.13$ & $1.51\pm.56$ & $6.35\pm1.16$ \\
 & NS & $.5$ & $.361\pm.004$ & $1.09\pm.76$ & $3.63\pm1.90$ & $1.54\pm.63$ & $6.58\pm1.01$ \\
 & EO & $5$ & $.367\pm.004$ & $1.07\pm.72$ & $3.14\pm2.16$ & $1.54\pm.56$ & $6.47\pm1.07$ \\
\midrule
B & SAFER & $5$ & $.381\pm.006$ & $1.14\pm1.06$ & $2.88\pm1.88$ & $1.25\pm.73$ & $4.90\pm1.28$ \\
 & NS & $1$ & $.379\pm.008$ & $1.23\pm.92$ & $3.30\pm2.48$ & $1.49\pm.81$ & $5.63\pm1.05$ \\
 & EO & $10$ & $.381\pm.006$ & $1.17\pm.86$ & $3.25\pm2.24$ & $1.44\pm.54$ & $5.98\pm1.08$ \\
\midrule
C & SAFER & $15$ & $.457\pm.031$ & $1.41\pm.50$ & $2.51\pm1.62$ & $.79\pm.76$ & $2.27\pm1.29$ \\
 & NS & $1.25$ & $.467\pm.066$ & $1.06\pm.84$ & $3.32\pm2.24$ & $1.84\pm1.26$ & $4.65\pm1.29$ \\
 & EO & $50$ & $.466\pm.011$ & $1.04\pm.91$ & $2.61\pm1.73$ & $1.08\pm.83$ & $4.29\pm.96$ \\
\bottomrule
\end{tabular}}
\end{table}

\begin{table}[p]
\centering\footnotesize
\caption{Complete $\lambda$ sweeps on the controlled benchmark: test
log-loss and mean residual certified excess per evidence stratum in
percentage points (mean $\pm$ standard deviation over 10 runs; lower is
better).  NS is NoSelectivity and EO is EO-Penalty.  Rows marked
$\dagger$ have mean test log-loss at or above $0.5$: these models
approach the constant predictor (log-loss $0.693$), and their small
excess values come from degenerate predictions rather than from fairer
ones.  The matched levels of Table~\ref{tab:synthetic-results} are
drawn from the unmarked rows.}
\label{tab:synthetic-sweeps}
\begin{tabular}{llcccccc}
\toprule
arm & $\lambda$ & test log-loss & null & weak & boundary & strong \\
\midrule
ERM & $0$ & $.356\pm.004$ & $.89\pm1.03$ & $3.49\pm2.69$ & $2.01\pm.74$ & $7.42\pm1.30$ \\
\midrule
SAFER & $0.5$ & $.357\pm.003$ & $1.08\pm.94$ & $4.02\pm2.43$ & $1.91\pm.84$ & $6.95\pm1.21$ \\
SAFER & $1$ & $.358\pm.003$ & $1.13\pm.79$ & $3.55\pm2.27$ & $1.76\pm.62$ & $6.77\pm1.38$ \\
SAFER & $1.5$ & $.360\pm.003$ & $1.19\pm.80$ & $3.16\pm2.12$ & $1.63\pm.49$ & $6.74\pm1.29$ \\
SAFER & $2$ & $.362\pm.003$ & $1.09\pm.77$ & $3.35\pm2.13$ & $1.51\pm.56$ & $6.35\pm1.16$ \\
SAFER & $2.5$ & $.365\pm.004$ & $1.02\pm.73$ & $3.28\pm2.07$ & $1.32\pm.65$ & $6.15\pm1.22$ \\
SAFER & $5$ & $.381\pm.006$ & $1.14\pm1.06$ & $2.88\pm1.88$ & $1.25\pm.73$ & $4.90\pm1.28$ \\
SAFER & $10$ & $.423\pm.018$ & $1.13\pm.67$ & $2.53\pm1.53$ & $.71\pm.59$ & $2.90\pm1.25$ \\
SAFER & $15$ & $.457\pm.031$ & $1.41\pm.50$ & $2.51\pm1.62$ & $.79\pm.76$ & $2.27\pm1.29$ \\
SAFER & $20$ & $.487\pm.037$ & $1.36\pm1.19$ & $2.87\pm1.98$ & $.63\pm.81$ & $2.01\pm1.58$ \\
SAFER & $30$ & $.539\pm.018$$^\dagger$ & $.75\pm1.04$ & $3.13\pm3.91$ & $.89\pm.86$ & $1.63\pm1.24$ \\
SAFER & $40$ & $.581\pm.021$$^\dagger$ & $.04\pm.07$ & $2.68\pm3.33$ & $.50\pm.75$ & $.31\pm.29$ \\
SAFER & $50$ & $.620\pm.024$$^\dagger$ & $.54\pm.96$ & $3.77\pm2.72$ & $.53\pm.73$ & $.43\pm.74$ \\
SAFER & $70$ & $.665\pm.013$$^\dagger$ & $.50\pm.82$ & $3.33\pm3.61$ & $.42\pm.77$ & $.14\pm.24$ \\
SAFER & $100$ & $.684\pm.004$$^\dagger$ & $.90\pm1.02$ & $2.62\pm2.36$ & $.24\pm.39$ & $.20\pm.23$ \\
SAFER & $150$ & $.690\pm.002$$^\dagger$ & $.20\pm.32$ & $.30\pm.52$ & $.12\pm.22$ & $.02\pm.06$ \\
SAFER & $200$ & $.692\pm.001$$^\dagger$ & $.03\pm.10$ & $.04\pm.12$ & $.01\pm.03$ & $.00\pm.00$ \\
SAFER & $300$ & $.694\pm.002$$^\dagger$ & $.00\pm.00$ & $.00\pm.00$ & $.00\pm.00$ & $.00\pm.00$ \\
SAFER & $500$ & $.696\pm.003$$^\dagger$ & $.00\pm.00$ & $.00\pm.00$ & $.00\pm.00$ & $.00\pm.00$ \\
\midrule
NS & $0.25$ & $.357\pm.004$ & $1.09\pm.88$ & $4.15\pm2.69$ & $1.83\pm.64$ & $6.80\pm1.10$ \\
NS & $0.5$ & $.361\pm.004$ & $1.09\pm.76$ & $3.63\pm1.90$ & $1.54\pm.63$ & $6.58\pm1.01$ \\
NS & $0.75$ & $.366\pm.005$ & $1.05\pm.68$ & $3.41\pm2.30$ & $1.38\pm.78$ & $6.15\pm1.26$ \\
NS & $1$ & $.379\pm.008$ & $1.23\pm.92$ & $3.30\pm2.48$ & $1.49\pm.81$ & $5.63\pm1.05$ \\
NS & $1.25$ & $.467\pm.066$ & $1.06\pm.84$ & $3.32\pm2.24$ & $1.84\pm1.26$ & $4.65\pm1.29$ \\
NS & $1.5$ & $.528\pm.073$$^\dagger$ & $.57\pm.74$ & $4.96\pm4.86$ & $1.73\pm1.21$ & $3.18\pm1.93$ \\
NS & $1.75$ & $.562\pm.077$$^\dagger$ & $.58\pm.58$ & $4.60\pm4.17$ & $.82\pm.94$ & $1.10\pm1.51$ \\
NS & $2$ & $.586\pm.071$$^\dagger$ & $.86\pm.90$ & $2.71\pm2.86$ & $.55\pm.43$ & $.95\pm1.24$ \\
NS & $2.25$ & $.608\pm.060$$^\dagger$ & $.48\pm.62$ & $1.90\pm2.28$ & $.40\pm.47$ & $.59\pm.78$ \\
NS & $2.5$ & $.627\pm.051$$^\dagger$ & $.36\pm.53$ & $1.32\pm1.70$ & $.30\pm.38$ & $.43\pm.51$ \\
NS & $2.75$ & $.643\pm.043$$^\dagger$ & $.30\pm.41$ & $1.15\pm1.74$ & $.13\pm.15$ & $.23\pm.38$ \\
NS & $3$ & $.653\pm.039$$^\dagger$ & $.37\pm.46$ & $.62\pm.85$ & $.18\pm.24$ & $.21\pm.28$ \\
NS & $3.5$ & $.671\pm.018$$^\dagger$ & $.14\pm.26$ & $.24\pm.55$ & $.04\pm.07$ & $.09\pm.22$ \\
NS & $3.75$ & $.677\pm.009$$^\dagger$ & $.14\pm.34$ & $.17\pm.45$ & $.02\pm.05$ & $.07\pm.20$ \\
NS & $4$ & $.681\pm.006$$^\dagger$ & $.09\pm.26$ & $.13\pm.39$ & $.01\pm.02$ & $.03\pm.08$ \\
NS & $4.5$ & $.685\pm.004$$^\dagger$ & $.07\pm.21$ & $.08\pm.23$ & $.00\pm.00$ & $.00\pm.00$ \\
NS & $5$ & $.688\pm.003$$^\dagger$ & $.03\pm.11$ & $.04\pm.12$ & $.02\pm.05$ & $.00\pm.00$ \\
NS & $5.4$ & $.689\pm.002$$^\dagger$ & $.02\pm.06$ & $.02\pm.07$ & $.01\pm.04$ & $.00\pm.00$ \\
\midrule
EO & $5$ & $.367\pm.004$ & $1.07\pm.72$ & $3.14\pm2.16$ & $1.54\pm.56$ & $6.47\pm1.07$ \\
EO & $10$ & $.381\pm.006$ & $1.17\pm.86$ & $3.25\pm2.24$ & $1.44\pm.54$ & $5.98\pm1.08$ \\
EO & $20$ & $.409\pm.009$ & $1.13\pm1.01$ & $2.87\pm2.09$ & $1.36\pm.78$ & $5.06\pm.91$ \\
EO & $50$ & $.466\pm.011$ & $1.04\pm.91$ & $2.61\pm1.73$ & $1.08\pm.83$ & $4.29\pm.96$ \\
EO & $60$ & $.479\pm.010$ & $1.03\pm.91$ & $2.57\pm1.73$ & $1.04\pm.76$ & $4.20\pm1.03$ \\
EO & $200$ & $.561\pm.011$$^\dagger$ & $1.20\pm.77$ & $3.27\pm1.91$ & $1.13\pm.93$ & $4.37\pm1.23$ \\
EO & $600$ & $.633\pm.011$$^\dagger$ & $.61\pm.76$ & $2.56\pm1.78$ & $1.23\pm.79$ & $3.89\pm1.94$ \\
EO & $1000$ & $.660\pm.008$$^\dagger$ & $.54\pm.54$ & $3.83\pm2.85$ & $.93\pm.76$ & $3.22\pm2.31$ \\
EO & $1500$ & $.674\pm.006$$^\dagger$ & $.68\pm.56$ & $3.20\pm1.59$ & $.77\pm.86$ & $2.20\pm1.72$ \\
EO & $2000$ & $.681\pm.004$$^\dagger$ & $.49\pm.67$ & $3.05\pm3.95$ & $.61\pm.66$ & $1.74\pm1.42$ \\
EO & $2500$ & $.684\pm.003$$^\dagger$ & $.42\pm.76$ & $3.13\pm3.28$ & $.55\pm.90$ & $1.41\pm1.08$ \\
EO & $5000$ & $.690\pm.002$$^\dagger$ & $.41\pm.67$ & $1.77\pm2.13$ & $.50\pm.65$ & $.35\pm.52$ \\
EO & $7500$ & $.691\pm.001$$^\dagger$ & $.39\pm.71$ & $2.03\pm2.85$ & $.31\pm.46$ & $.12\pm.22$ \\
EO & $10000$ & $.692\pm.001$$^\dagger$ & $.27\pm.52$ & $1.27\pm1.66$ & $.42\pm.75$ & $.06\pm.13$ \\
\bottomrule
\end{tabular}
\end{table}

\begin{table}[t]
\centering\footnotesize
\caption{Raw EO gap $|\phi_S|$ on test and on training members, mean
over the cells of each evidence stratum, for the non-degenerate points of
the sweeps (test log-loss below $0.5$; mean $\pm$ standard deviation over
10 runs).  Train and test values come from the same model.  A training
gap that falls much faster than the test gap on the sparse strata (weak,
boundary) indicates fitting of training-set noise.}
\label{tab:synthetic-absphi}
\resizebox{\linewidth}{!}{%
\begin{tabular}{llccccccccc}
\toprule
 & & & \multicolumn{4}{c}{test $|\phi_S|$} & \multicolumn{4}{c}{train $|\phi_S|$} \\
\cmidrule(lr){4-7}\cmidrule(lr){8-11}
arm & $\lambda$ & test log-loss & null & weak & boundary & strong & null & weak & boundary & strong \\
\midrule
ERM & $0$ & $.356\pm.004$ & $.048\pm.015$ & $.200\pm.027$ & $.126\pm.027$ & $.137\pm.013$ & $.054\pm.009$ & $.185\pm.031$ & $.124\pm.022$ & $.134\pm.017$ \\
\midrule
SAFER & $0.5$ & $.357\pm.003$ & $.050\pm.015$ & $.201\pm.028$ & $.124\pm.030$ & $.131\pm.012$ & $.050\pm.011$ & $.175\pm.026$ & $.111\pm.021$ & $.125\pm.015$ \\
SAFER & $1$ & $.358\pm.003$ & $.050\pm.014$ & $.187\pm.027$ & $.122\pm.029$ & $.129\pm.014$ & $.046\pm.011$ & $.170\pm.027$ & $.100\pm.021$ & $.116\pm.015$ \\
SAFER & $1.5$ & $.360\pm.003$ & $.049\pm.014$ & $.182\pm.025$ & $.118\pm.028$ & $.128\pm.013$ & $.045\pm.009$ & $.169\pm.027$ & $.095\pm.019$ & $.110\pm.014$ \\
SAFER & $2$ & $.362\pm.003$ & $.047\pm.013$ & $.185\pm.024$ & $.117\pm.027$ & $.124\pm.012$ & $.043\pm.009$ & $.162\pm.026$ & $.092\pm.016$ & $.106\pm.012$ \\
SAFER & $2.5$ & $.365\pm.004$ & $.045\pm.011$ & $.182\pm.025$ & $.115\pm.026$ & $.121\pm.013$ & $.040\pm.008$ & $.155\pm.020$ & $.087\pm.013$ & $.103\pm.012$ \\
SAFER & $5$ & $.381\pm.006$ & $.041\pm.009$ & $.170\pm.023$ & $.105\pm.033$ & $.105\pm.015$ & $.038\pm.009$ & $.139\pm.018$ & $.071\pm.007$ & $.085\pm.014$ \\
SAFER & $10$ & $.423\pm.018$ & $.033\pm.007$ & $.143\pm.027$ & $.084\pm.026$ & $.077\pm.018$ & $.031\pm.007$ & $.099\pm.016$ & $.054\pm.010$ & $.053\pm.008$ \\
SAFER & $15$ & $.457\pm.031$ & $.035\pm.013$ & $.129\pm.031$ & $.080\pm.034$ & $.068\pm.022$ & $.027\pm.006$ & $.087\pm.025$ & $.044\pm.015$ & $.036\pm.007$ \\
SAFER & $20$ & $.487\pm.037$ & $.041\pm.023$ & $.129\pm.047$ & $.076\pm.038$ & $.065\pm.027$ & $.031\pm.014$ & $.095\pm.035$ & $.041\pm.016$ & $.034\pm.011$ \\
\midrule
NS & $0.25$ & $.357\pm.004$ & $.050\pm.015$ & $.194\pm.033$ & $.123\pm.026$ & $.129\pm.011$ & $.046\pm.014$ & $.147\pm.027$ & $.111\pm.025$ & $.125\pm.014$ \\
NS & $0.5$ & $.361\pm.004$ & $.048\pm.014$ & $.191\pm.031$ & $.119\pm.028$ & $.127\pm.010$ & $.040\pm.010$ & $.128\pm.021$ & $.102\pm.024$ & $.117\pm.014$ \\
NS & $0.75$ & $.366\pm.005$ & $.044\pm.010$ & $.183\pm.028$ & $.115\pm.028$ & $.121\pm.013$ & $.038\pm.009$ & $.121\pm.019$ & $.094\pm.024$ & $.113\pm.015$ \\
NS & $1$ & $.379\pm.008$ & $.044\pm.011$ & $.171\pm.027$ & $.113\pm.030$ & $.114\pm.011$ & $.035\pm.011$ & $.106\pm.014$ & $.080\pm.015$ & $.105\pm.018$ \\
NS & $1.25$ & $.467\pm.066$ & $.053\pm.018$ & $.172\pm.039$ & $.120\pm.041$ & $.106\pm.017$ & $.037\pm.013$ & $.094\pm.017$ & $.080\pm.023$ & $.102\pm.029$ \\
\midrule
EO & $5$ & $.367\pm.004$ & $.048\pm.013$ & $.182\pm.027$ & $.119\pm.030$ & $.125\pm.011$ & $.045\pm.012$ & $.127\pm.021$ & $.099\pm.018$ & $.118\pm.016$ \\
EO & $10$ & $.381\pm.006$ & $.046\pm.011$ & $.182\pm.026$ & $.114\pm.027$ & $.119\pm.012$ & $.042\pm.009$ & $.108\pm.015$ & $.090\pm.015$ & $.109\pm.017$ \\
EO & $20$ & $.409\pm.009$ & $.043\pm.011$ & $.171\pm.021$ & $.108\pm.029$ & $.108\pm.010$ & $.039\pm.008$ & $.102\pm.010$ & $.074\pm.012$ & $.099\pm.018$ \\
EO & $50$ & $.466\pm.011$ & $.040\pm.011$ & $.163\pm.024$ & $.099\pm.030$ & $.098\pm.010$ & $.036\pm.009$ & $.095\pm.013$ & $.064\pm.012$ & $.088\pm.022$ \\
EO & $60$ & $.479\pm.010$ & $.041\pm.011$ & $.157\pm.020$ & $.099\pm.030$ & $.097\pm.011$ & $.036\pm.011$ & $.096\pm.014$ & $.064\pm.012$ & $.087\pm.021$ \\
\bottomrule
\end{tabular}}
\end{table}

\begin{table}[h]
\centering\small
\caption{Size sweep: ERM models on a variant of the benchmark with one
reference cell ($n=150{,}000$, true-positive rate $0.85$) and six probe
cells per design size $m$, all with planted gap $-0.15$.  For each $m$,
the mean certification statistic $\sqrt{\Lambda_S}$ over the probe
cells and runs, computed on hard training predictions of the final ERM
model, and the fraction of probe cells certified
($\sqrt{\Lambda_S}>1.96$) as mean $\pm$ standard deviation over 10 runs.
At $m=5$, two of the 60 cells had fewer than two positives and were
skipped.}
\label{tab:synthetic-size}
\resizebox{\linewidth}{!}{%
\begin{tabular}{lccccccccc}
\toprule
$m$ & $5$ & $10$ & $15$ & $20$ & $30$ & $50$ & $80$ & $120$ & $150$ \\
\midrule
mean $\sqrt{\Lambda_S}$ & $1.03$ & $1.29$ & $1.55$ & $1.49$ & $1.96$ & $2.15$ & $2.99$ & $3.57$ & $4.11$ \\
certified fraction & $.09\pm.09$ & $.23\pm.16$ & $.33\pm.18$ & $.27\pm.18$ & $.52\pm.15$ & $.58\pm.24$ & $.85\pm.18$ & $.95\pm.08$ & $.97\pm.07$ \\
cells (all runs) & $58$ & $60$ & $60$ & $60$ & $60$ & $60$ & $60$ & $60$ & $60$ \\
\bottomrule
\end{tabular}}
\end{table}

\section{Ablation study details}
\label{app:ablation-details}

The arm definitions and detailed numbers for the mechanism
ablations of Section~\ref{sec:exp-ablation} are given below.

\subsection{Arm definitions}

The four arms are evaluated on the three stress variants of
Table~\ref{tab:datasets} under the protocols of
Section~\ref{sec:exp-protocol}:
\begin{itemize}[leftmargin=*,noitemsep,topsep=2pt]
\item \textbf{SAFER}: full method, all three components active.
\item \textbf{NoThreshSmooth}: \(\tau\) disabled (raw
  probabilities feed the soft counts).
\item \textbf{NoSelectivity}: the threshold term
  \(z_\alpha\widehat{SE}_{\mathrm{eff},S}\) is removed from the
  softplus argument, leaving
  \(\sqrt{\widehat\phi^\tau_S{}^2+\varepsilon^2}\) alone.
\item \textbf{NoDenseTail}: \(w_{\mathrm{tail}}=0\).
\end{itemize}
Apart from the toggled knob, every other element of the loss,
optimizer, and protocol is identical across arms.

\subsection{Per-mechanism detailed numbers}

\paragraph{Threshold smoothing.}
On Adult L4 10/90 (10 splits$\times$3 seeds), reaching
\(MSEE\!\approx\!10^{-3}\) requires \(\lambda\!=\!1000\) at log-loss
\(\approx 0.57\) for NoThreshSmooth, versus \(\lambda\!=\!200\) at
log-loss \(\approx 0.43\) for SAFER. At matched fairness, the
difference is roughly \(0.14\) nats.  On COMPAS L3 2080, the gap is
\(0.01\)--\(0.02\) nats at matched MSEE through the mid-range
(Figure~\ref{fig:ablation-compas-german}, center).

\paragraph{Dense tail.}
On COMPAS L3 2080, at \(\lambda\!\le\!2\) the four arms are within the
\(\pm\)SE bands of one another.  From \(\lambda\!=\!5\) onward,
NoDenseTail plateaus at \(MSEE\!\approx\!0.06\); the gap to SAFER
widens monotonically and reaches roughly ten standard errors at
\(\lambda\!=\!100\) (SAFER \(0.018\!\pm\!0.001\) versus NoDenseTail
\(0.060\!\pm\!0.004\)).  On Adult L4 1090 the truncation has a different scale but the
predicted form: NoDenseTail
stays close to ERM MSEE (\(\approx 0.02\)) for almost every
\(\lambda\), and only at the largest tested \(\lambda\) does it
move toward smaller MSEE, where it is Pareto-dominated by SAFER.
On German L3 50/50 NoDenseTail tracks SAFER, because the
significance gate already keeps the hinge inactive
(Proposition~\ref{cor:safety}).

\paragraph{Gate threshold (NoSelectivity on German).}
SAFER preserves ERM-level log-loss (\(\approx 0.53\)) up to
\(\lambda\!=\!20\) and degrades only mildly (\(0.59\) at
\(\lambda\!=\!100\)).  NoSelectivity climbs from log-loss \(0.53\) at
\(\lambda\!=\!1\) through \(0.60\) at \(\lambda\!=\!5\) and reaches
\(0.69\) by \(\lambda\!=\!20\); the error bars on its MSEE become very
large in this range.

\subsection{NoThreshSmooth on German: no departure from ERM}

NoThreshSmooth on German exhibits a different and benign failure mode
from NoSelectivity: the raw soft-count gradient is too small to move
the model on this small dataset, so the optimizer never leaves ERM
regardless of \(\lambda\).  This is consistent with
Proposition~\ref{prop:raw-inconsistency}: when subgroup distributions
are tightly clustered well above (or below) the decision boundary, raw
probabilities and hard decisions disagree but the threshold-smoothing
step that would expose the hard-decision gap is absent.  This safe behavior is therefore a side effect of the small
gradient and is not guaranteed by the method.

\subsection{Selective-gating: what did not appear}
\label{app:selective-gating-real}

The hypothesis for the selective-gating regime, stated in
Appendix~\ref{app:synthetic}, is that the gate threshold gives
SAFER an advantage over an ungated alternative when active and
inactive subgroups coexist.  The Adult L4 1090 and COMPAS L3 2080
results in Figure~\ref{fig:ablation-compas-german} (left
and center) do not exhibit this behavior: NoSelectivity tracks SAFER
up to $\lambda=1$ on Adult, where it then loses utility, and across
the sweep on COMPAS.  We interpret
this as evidence that the gate's empirical role is asymmetric on
these two datasets: it suppresses harm in the safety regime, while
the evidence for its selective-regime benefit is directional and comes
from the controlled benchmark of Appendix~\ref{app:synthetic}, where the
evidence strata are isolated.

\subsection{Cross-regime ablation map}
\label{app:cross-regime}

\begin{table}[h]
\centering
\small
\caption{Empirical effect of removing each component on each stress
variant.  ``\(\approx\!0\)'' indicates indistinguishable from full
SAFER within \(\pm\)SE.  ``\(\downarrow\!\downarrow\)'' indicates
Pareto-dominated by SAFER across the regularization sweep.
``\(\downarrow\)'' indicates a Pareto gap that opens at high
\(\lambda\).  ``harm'' indicates that the ablation degrades utility
without any certified-fairness benefit.  ``n/a'' indicates the
ablation was not run rigorously at the matched protocol.}
\label{tab:component-regime}
\begin{tabular}{lccc}
\toprule
Ablation & Adult L4 1090 & COMPAS L3 2080 & German L3 5050 \\
\midrule
NoThreshSmooth (Theorem~\ref{thm:thresh-consistency})
                  & \(\downarrow\!\downarrow\) & \(\downarrow\!\downarrow\)
                  & \(\approx\!0\) (no traction) \\
NoDenseTail (Eq.~\eqref{eq:hinge-suppression})
                  & \(\downarrow\) at high \(\lambda\) & \(\downarrow\) at high \(\lambda\)
                  & \(\approx\!0\) \\
NoSelectivity (Proposition~\ref{cor:safety})
                  & \(\approx\!0\), harm at high \(\lambda\) & \(\approx\!0\) & harm \\
\bottomrule
\end{tabular}
\end{table}

The Eq.~\eqref{eq:hinge-suppression} mechanism predicts that
NoDenseTail under-reaches SAFER at high \(\lambda\) on
certified-violation datasets, and we observe this on both Adult L4
1090 and COMPAS L3 2080.  The shape of the truncation differs:
COMPAS shows a clean plateau at \(MSEE\!\approx\!0.06\) from
\(\lambda\!=\!5\) onward, while Adult shows NoDenseTail stuck near
ERM MSEE for almost the entire \(\lambda\) sweep, escaping only at
the largest tested \(\lambda\) and remaining Pareto-dominated by
SAFER (Figure~\ref{fig:ablation-compas-german}, left).  Both shapes
are consistent with the same hinge-suppression argument; a more
controlled comparison of where each dataset's subgroup statistics
sit relative to the gate boundary would tighten the
quantitative match and is left to future work.

\section{Statistical Parity results}
\label{app:sp}

\begin{figure}[h]
\centering
\begin{subfigure}{0.48\linewidth}
    \centering
    \includegraphics[width=\linewidth]{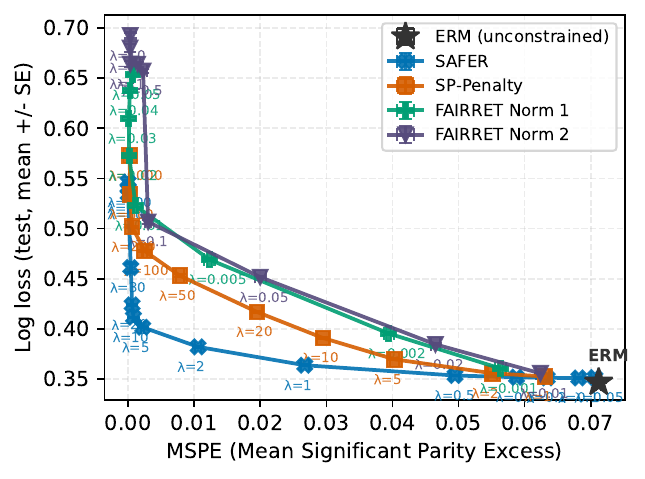}
    \caption{Adult SP Pareto frontier.}
    \label{fig:adult-sp-pareto}
\end{subfigure}
\begin{subfigure}{0.48\linewidth}
    \centering
    \includegraphics[width=\linewidth]{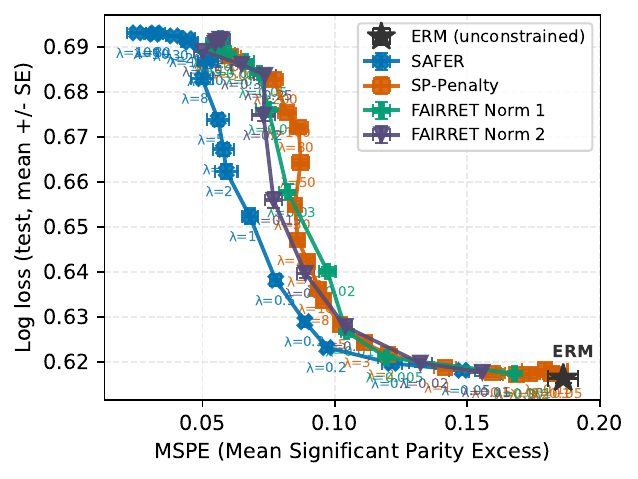}
    \caption{COMPAS SP Pareto frontier.}
    \label{fig:compas-sp-pareto}
\end{subfigure}
\begin{subfigure}{0.48\linewidth}
    \centering
    \includegraphics[width=\linewidth]{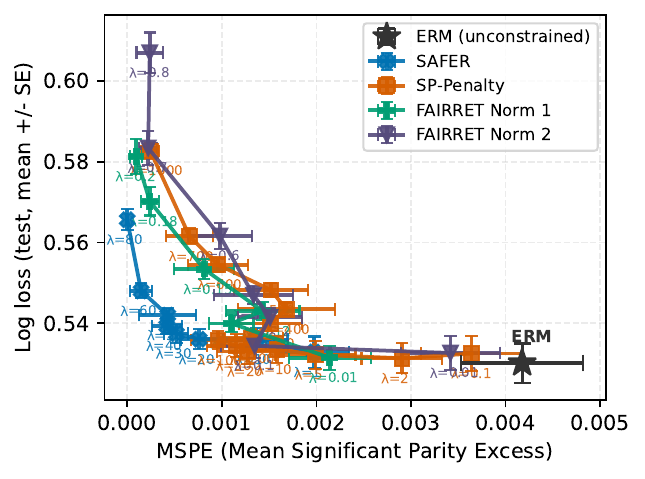}
    \caption{German Credit SP Pareto frontier.}
    \label{fig:german-sp-pareto}
\end{subfigure}
\caption{Pareto frontiers for Adult, COMPAS and German Credit under Statistical Parity.}
\label{fig:sp-pareto-frontiers}
\end{figure}

The main text focuses on EO because its positive-label conditioning makes the
sparse-cell issue most severe.  SP has larger effective subgroup sizes because
it conditions on all examples in a subgroup.  Figure~\ref{fig:sp-pareto-frontiers}
shows the Adult, COMPAS and German SP Pareto frontiers. The penalty
baseline, SP-Penalty, is the certainty-blind squared-disparity regularizer
applied to the SP gap. Results are consistent with the EO interpretation in
Section~\ref{sec:experiments}.

\section{Other instantiations}
\label{app:beyond-msee}
\subsection{Equalized Odds}

Section~\ref{sec:safer} shows that the SAFER framework extends beyond the
single-component SP/EO case to multi-component criteria, by summing
componentwise hinges.  We demonstrate this concretely for
\textbf{Equalized Odds}~\citep{hardt2016equality}, which requires simultaneous
parity of the true-positive rate (TPR) and the false-positive rate (FPR) across
subgroups:
\begin{align}
    EO(S) &= P(\hat{Y}{=}1 \mid Y{=}1, S) - P(\hat{Y}{=}1 \mid Y{=}1, \bar{S}), \\
    FP(S) &= P(\hat{Y}{=}1 \mid Y{=}0, S) - P(\hat{Y}{=}1 \mid Y{=}0, \bar{S}).
\end{align}
The TPR component is the Equal Opportunity criterion already treated in the main
paper; FPR parity is the complementary condition on negative-label examples.

\spara{FPR soft counts.}
Section~\ref{sec:safer} defines TPR soft counts by restricting to
positive-label examples.  The FPR component requires the analogous restriction
to negative-label examples:
\begin{align}
    \tilde{n}_{1,S}^{(-)} &= \sum_{i:\,y_i=0} \hat{y}_i \,\mathbf{1}_S(i),
    &
    \tilde{n}_{0,S}^{(-)} &= \sum_{i:\,y_i=0} (1-\hat{y}_i)\,\mathbf{1}_S(i).
\end{align}
These induce a differentiable FPR estimate
$\hat{f}_S = \tilde{n}_{1,S}^{(-)}\big/\!\left(\tilde{n}_{1,S}^{(-)}+\tilde{n}_{0,S}^{(-)}\right)$
and, via the identical two-sample proportion construction, a differentiable FPR
disparity $\widehat{FP}_S(\theta)=\hat{f}_S-\hat{f}_{\bar{S}}$ and LRT
statistic $T_S^{\mathrm{FPR}}(\theta)=\sqrt{\Lambda_S^{\mathrm{FPR}}}$ where
$\Lambda_S^{\mathrm{FPR}}$ is the log-likelihood ratio for the null of equal
FPR between $S$ and $\bar{S}$.

\spara{Combined SAFER objective.}
The Equalized Odds variant penalizes both components independently:
\begin{equation}
\label{eq:safer-eqodds}
    \mathcal{L}_{\mathrm{SAFER}}^{\mathrm{EqOdds}}(\theta)
    =
    \mathcal{L}_{\mathrm{pred}}(\theta)
    +
    \frac{\lambda}{|\mathfrak{S}|}
    \sum_{S\in\mathfrak{S}}
    \Bigl[
        \mathcal L_{\mathrm{dir},S}^{\mathrm{TPR}}(\theta)
        +
        \mathcal L_{\mathrm{dir},S}^{\mathrm{FPR}}(\theta)
    \Bigr].
\end{equation}
Each component uses the same direct significant-excess form as
Eq.~\eqref{eq:direct-surrogate}, but with soft counts restricted to the
corresponding label stratum.  The significance gate and the safety argument apply
componentwise: a subgroup whose TPR and FPR components both lie below the
gate threshold has only the
exponentially small below-threshold gradient described by
Eq.~\eqref{eq:hinge-suppression}, plus any dense-tail residual.

\spara{Effective sample sizes.}
For EqOdds, the binding constraint is the smaller of the two label-conditioned
counts:
\[
    m_S^{\mathrm{EqOdds}} = \min\!\bigl(m_S^{(+)},\, m_S^{(-)}\bigr),
    \quad
    m_S^{(+)} = |\{i\in S : y_i=1\}|,
    \quad
    m_S^{(-)} = |\{i\in S : y_i=0\}|.
\]
Whenever $\hat\rho \ll 1$ (the typical regime for sparse intersectional
subgroups), the positive-label count $m_S^{(+)}\approx n_S\hat\rho_S$ is the
binding term; when negative labels are rarer, the FPR component is binding
instead.

\spara{Metric.}
Specializing Definition~\ref{def:significant-excess} to the two components
yields the combined certified-excess metric:
\begin{equation}
\label{eq:msee-eqodds}
    \mathrm{MSE}^{\mathrm{EqOdds}}_\alpha(\theta)
    =
    \frac{1}{|\mathfrak{S}|}
    \sum_{S\in\mathfrak{S}}
    \bigl[
        \mathcal{E}^{\mathrm{TPR}}_{\alpha}(S) + \mathcal{E}^{\mathrm{FPR}}_{\alpha}(S)
    \bigr],
\end{equation}
where each component is the significant excess of
Definition~\ref{def:significant-excess} applied to the respective label
stratum.  This metric is zero if and only if no subgroup has a certified
violation in either the TPR or FPR component, inheriting property~1 of the
significant-excess family.

\subsection{Disparate Impact instantiation}

Disparate Impact (DI)~\citep{feldman2015certifying} measures the ratio of
positive-prediction rates across subgroups:
\[
    DI(S) = \frac{P(\hat{Y}{=}1 \mid S)}{P(\hat{Y}{=}1 \mid \bar{S})}.
\]
We use DI in its \emph{parity} formulation, which tests whether $DI(S)=1$,
rather than the thresholded $0.8$-rule used in legal practice.  The parity null
$DI(S)=1$ is equivalent to $P(\hat{Y}{=}1\mid S)=P(\hat{Y}{=}1\mid\bar{S})$,
which is exactly the null of Statistical Parity.  Consequently,
test statistic and the two-sample LRT statistic for DI parity are identical
to the one already defined for SP.
Therefore, the SAFER training objective for DI is the same as the SP
variant of Eq.~\eqref{eq:safer-objective}.

Although the training objective is shared with SP, the significant-excess metric
can be expressed in log-ratio units, by defining the log-DI functional
$\phi^{DI}(S)=\log DI(S)=\log p_S - \log p_{\bar{S}}$. The significant-excess is derived by following Definition \ref{def:significant-excess} given a confidence interval $[\ell_\alpha(S),\, u_\alpha(S)]$ from a statistical test on $\phi^{DI}(S)$.

\section{Proofs for Section~\ref{sec:safer}}
\label{app:proofs-v2}

Throughout this appendix, we write
\(\sigma_\tau(u)=\sigma(u/\tau)\), where \(\sigma\) is the logistic
sigmoid, and
\(\operatorname{softplus}_\tau(u)=\tau\log(1+e^{u/\tau})\).

\subsection{Basic sigmoid bounds}

\begin{lemma}[Sigmoid--step difference]
\label{lem:sigmoid-step}
Let \(\ \sigma \) be the logistic sigmoid and  \( \sigma_\tau(u)=\sigma(u/\tau) \). For every \(u\in\mathbb{R}\) and \(\tau>0\),
\[
    \big|\sigma_\tau(u)-\mathbf{1}\{u>0\}\big|
    =
    \sigma(-|u|/\tau).
\]
\end{lemma}
\begin{proof}
If \(u>0\), the difference is
\(1-\sigma(u/\tau)=\sigma(-u/\tau)\).  If \(u<0\), it is
\(\sigma(u/\tau)=\sigma(-|u|/\tau)\).  At \(u=0\), both sides are
\(1/2\).
\end{proof}

\begin{lemma}[Integral and derivative bounds]
\label{lem:sigmoid-tools}
For every \(\tau>0\),
\[
    \int_{-\infty}^{\infty}\sigma(-|v|/\tau)\,dv=2\tau\log 2,
    \qquad
    \sigma'(z)\le e^{-|z|}\quad\forall z\in\mathbb{R}.
\]
\end{lemma}
\begin{proof}
The integral follows by the change of variables \(w=v/\tau\):
\[
    \int_{-\infty}^{\infty}\sigma(-|v|/\tau)\,dv
    =
    2\tau\int_0^\infty \frac{e^{-w}}{1+e^{-w}}\,dw
    =
    2\tau\log 2.
\]
For the derivative bound, use
\(\sigma'(z)=\sigma(z)(1-\sigma(z))\).  If \(z\ge0\), then
\(1-\sigma(z)=\sigma(-z)\le e^{-z}\). Symmetry gives the case \(z<0\).
\end{proof}

\subsection{Proof of Theorem~\ref{thm:thresh-consistency}}

Let \(F_S\) be the conditional distribution of \(f_\theta(X)\) on the
conditioning population \(\Pi_S\), and define
\[
    p_S^\tau
    =
    \mathbb{E}_{F_S}\!\left[\sigma_\tau(f_\theta(X)-\tfrac12)\right],
    \qquad
    p_S^{\mathrm{hard}}
    =
    \mathbb{E}_{F_S}\!\left[\mathbf{1}\{f_\theta(X)>\tfrac12\}\right].
\]
By Lemma~\ref{lem:sigmoid-step},
\[
    |p_S^\tau-p_S^{\mathrm{hard}}|
    \le
    \int \sigma(-|u-\tfrac12|/\tau)\,dF_S(u).
\]
Let \(N=\{u:|u-\tfrac12|\le\delta\}\) be the interval on which, by
assumption, \(F_S\) has a density bounded by \(L\), so that
\(dF_S(u)\le L\,du\) on \(N\).  We split the integral at \(N\).  On \(N\),
substituting \(v=u-\tfrac12\), extending the range of integration to the
real line and using Lemma~\ref{lem:sigmoid-tools},
\[
    \int_{N}\sigma(-|u-\tfrac12|/\tau)\,dF_S(u)
    \le
    L\int_{-\infty}^{\infty}\sigma(-|v|/\tau)\,dv
    =
    2L\tau\log 2 .
\]
Outside \(N\) we have \(|u-\tfrac12|>\delta\), so
\(\sigma(-|u-\tfrac12|/\tau)\le\sigma(-\delta/\tau)\le e^{-\delta/\tau}\),
and the mass of \(F_S\) outside \(N\) is at most one; hence
\[
    \int_{N^{c}}\sigma(-|u-\tfrac12|/\tau)\,dF_S(u)
    \le
    e^{-\delta/\tau}.
\]
Adding the two parts,
\[
    |p_S^\tau-p_S^{\mathrm{hard}}|
    \le
    2L\tau\log 2 + e^{-\delta/\tau}.
\]
The same bound holds for \(\bar S\). Since
\(\phi_S^\bullet=p_S^\bullet-p_{\bar S}^\bullet\), the triangle
inequality gives:
\[
    |\phi_S^\tau-\phi_S^{\mathrm{hard}}|
    \le
    |p_S^\tau-p_S^{\mathrm{hard}}|
    +
    |p_{\bar S}^\tau-p_{\bar S}^{\mathrm{hard}}|
    \le
    4L\tau\log 2 {}+ 2e^{-\delta/\tau}.
\]
The remainder is also linear in \(\tau\): since \(e^{-x}\le 1/(ex)\)
for \(x>0\), we have \(2e^{-\delta/\tau}\le 2\tau/(e\delta)\), so the whole
bound is at most \((4L\log 2+2/(e\delta))\,\tau\).  Outside \(N\) the score
distribution is arbitrary; in particular it may have atoms at \(0\) and
\(1\).
\(\square\)

\subsection{Proof of Proposition~\ref{prop:raw-inconsistency}}

Fix \(\eta>0\).  Choose \(\kappa<\eta/2\) and let \(F_S\) concentrate
smoothly in a small interval around \(1/2+\kappa\), while \(F_{\bar S}\)
concentrates smoothly in a small interval around \(1/2-\kappa\).  Then
\[
    \phi_S^{\mathrm{raw}} \approx 2\kappa,
    \qquad
    \phi_S^{\mathrm{hard}}\approx 1,
\]
so
\[
    |\phi_S^{\mathrm{raw}}-\phi_S^{\mathrm{hard}}| 
    \ge 1-\eta
\]
for sufficiently narrow intervals.  The distributions may be chosen with
continuous densities bounded by a finite constant.

To see that no constant rescaling works uniformly, compare this construction
with a second one in which both groups are well above the threshold, with
mean scores differing by \(\Delta>0\).  Then
\(\phi_S^{\mathrm{raw}}\approx\Delta\) while
\(\phi_S^{\mathrm{hard}}=0\).  Matching the first construction as
\(\kappa\to0\) requires a rescaling constant of order \(1/(2\kappa)\), which
then makes the rescaled raw disparity in the second construction diverge
instead of matching zero.  Hence a single constant cannot align raw soft
counts with hard-decision disparities across both regimes. \(\square\)

\subsection{Gradient localization at the decision boundary}

The following result, deferred from Section~\ref{sec:safer-thresholding},
shows that threshold smoothing localizes the fairness gradient to
examples near the decision boundary $\{x:f_\theta(x)=\frac12\}$.

\begin{theorem}[Gradient localization]
\label{thm:gradient-localization}
For SP and EO,
\begin{equation}
\label{eq:grad-localization-bound}
    \biggl|\frac{\partial\phi^\tau_S}{\partial f_\theta(x)}\biggr|
    \le
    \frac{c_S(x)}{\tau}\,
    \exp\!\biggl(-\frac{|f_\theta(x)-\frac12|}{\tau}\biggr),
\end{equation}
for a bounded subgroup-membership weight $c_S(x)$.  As $\tau\to0^+$,
the gradient concentrates on $\{x:f_\theta(x)=\frac12\}$.
\end{theorem}
\begin{proof}
For \(x\in\Pi_S\),
\[
    \frac{\partial p_S^\tau}{\partial f_\theta(x)}
    =
    \frac{1}{\tau}
    \sigma'\!\left((f_\theta(x)-\tfrac12)/\tau\right)
    c_S^+(x),
\]
where \(c_S^+(x)\) is the appropriate conditioning-population weight.  The
corresponding complement term has the same form with a possibly different
weight.  Therefore, for a bounded subgroup-membership weight \(c_S(x)\),
\[
    \Big|
    \frac{\partial\phi_S^\tau}{\partial f_\theta(x)}
    \Big|
    \le
    \frac{c_S(x)}{\tau}
    \sigma'\!\left((f_\theta(x)-\tfrac12)/\tau\right)
    \le
    \frac{c_S(x)}{\tau}
    e^{-|f_\theta(x)-1/2|/\tau},
\]
using Lemma~\ref{lem:sigmoid-tools}.  Finally,
\(\tau^{-1}\sigma'((u-1/2)/\tau)\) is an approximate identity that converges
weakly to a Dirac mass at \(u=1/2\), yielding the distributional limit.
\end{proof}

\subsection{Canonical surrogate approximation}
\label{sec:safer-direct-appendix}

The main text (Eq.~\eqref{eq:direct-surrogate}) uses an LRT-based
threshold.  Here we state and prove a metric-alignment bound for a
canonical version in which the surrogate and the metric share the population
Wald threshold $B_S$.  The bound shows that the surrogate \emph{shape} tracks
$MSEE^{\mathrm{pop}}_\alpha$, defined below.
This population comparison does not bound the discrepancy between the
implemented LRT-based training loss and the SAFT audit metric.

Let $B_S(\theta)=z_\alpha\,SE^\phi_S(\theta)$, where $SE^\phi_S$ is
evaluated with population cell probabilities.
Define
\begin{align}
\label{eq:canonical-msee}
    MSEE^{\mathrm{pop}}_\alpha(\theta)
    &=
    \frac{1}{|\mathfrak S|}
    \sum_{S\in\mathfrak S}
    \bigl[|\phi^{\mathrm{hard}}_S(\theta)|-B_S(\theta)\bigr]_+\,,
    \\[4pt]
\label{eq:canonical-surrogate}
    \mathcal L_{\mathrm{can}}(\theta)
    &=
    \frac{1}{|\mathfrak S|}
    \sum_{S\in\mathfrak S}
    \operatorname{softplus}_{\tau_s}\!\left(
        \sqrt{\phi^\tau_S(\theta)^2+\varepsilon^2}
        - B_S(\theta)
    \right).
\end{align}

\begin{theorem}[Canonical surrogate approximation]
\label{thm:surrogate-approx}
Under Theorem~\ref{thm:thresh-consistency}'s density assumption, with
the same $L$ and $\delta$, for
every $\varepsilon,\tau_s,\tau>0$,
\begin{equation}
\label{eq:approx-bound}
    \bigl|\mathcal L_{\mathrm{can}}(\theta)
    -MSEE^{\mathrm{pop}}_\alpha(\theta)\bigr|
    \le
    \varepsilon + \tau_s\log 2 + 4L\tau\log 2{}+2e^{-\delta/\tau}\,.
\end{equation}
\end{theorem}
\begin{proof}
Fix a subgroup \(S\), write \(B_S=B_S(\theta)\), and define
\[
    h_S
    =
    \operatorname{softplus}_{\tau_s}
    \left(\sqrt{(\phi_S^\tau)^2+\varepsilon^2}-B_S\right),
    \qquad
    m_S
    =
    \big[|\phi_S^{\mathrm{hard}}|-B_S\big]_+ .
\]
The elementary inequalities
\[
    \big|\sqrt{x^2+\varepsilon^2}-|x|\big|\le\varepsilon,
    \qquad
    0\le \operatorname{softplus}_{\tau_s}(u)-[u]_+\le\tau_s\log2,
\]
and the Lipschitz property of \(x\mapsto[x]_+\) imply
\[
    |h_S-m_S|
    \le
    \varepsilon+\tau_s\log2+
    \big||\phi_S^\tau|-|\phi_S^{\mathrm{hard}}|\big|.
\]
Since \(\big||a|-|b|\big|\le |a-b|\), Theorem~\ref{thm:thresh-consistency}
gives
\[
    |h_S-m_S|
    \le
    \varepsilon+\tau_s\log2+4L\tau\log2{}+2e^{-\delta/\tau} .
\]
Averaging this bound over \(S\in\mathfrak{S}\) proves
Eq.~\eqref{eq:approx-bound}.
\end{proof}

\subsection{Asymptotic equivalence of the soft-count LRT and the Wald statistic}
\label{app:property-iii}
Write $Q_i=\sigma_\tau(f_\theta(x_i)-\tfrac12)\in(0,1)$ for the soft
count of observation $i$, so that $\hat p_S$ and $\hat p_{\bar S}$ are
the means of the $Q_i$ over $\Pi_S$ and $\Pi_{\bar S}$, $\hat p$ is the
pooled mean, and $\tilde n_S=|\Pi_S|$, $\tilde n_{\bar S}=|\Pi_{\bar S}|$.
With $D(a\,\|\,b)=a\log\frac{a}{b}+(1-a)\log\frac{1-a}{1-b}$ the Bernoulli
divergence, Eq.~\eqref{eq:lrt-soft} reads
\[
    \Lambda_S
    =
    2\bigl[\tilde n_S\,D(\hat p_S\,\|\,\hat p)
    +\tilde n_{\bar S}\,D(\hat p_{\bar S}\,\|\,\hat p)\bigr].
\]
Set $\Delta=\hat p_S-\hat p_{\bar S}$ and $N=\tilde n_S+\tilde n_{\bar S}$,
so that $\hat p_S-\hat p=(\tilde n_{\bar S}/N)\,\Delta$ and
$\hat p_{\bar S}-\hat p=-(\tilde n_S/N)\,\Delta$.  Since
$D(a\,\|\,b)=\frac{(a-b)^2}{2b(1-b)}+O(|a-b|^3)$ for $b$ bounded away
from $0$ and $1$,
\[
    \Lambda_S
    =
    \frac{\Delta^2}{\hat p(1-\hat p)\,(1/\tilde n_S+1/\tilde n_{\bar S})}
    \,\bigl(1+O(|\Delta|)\bigr).
\]
The plug-in Wald variance of property (ii) satisfies
$(\widehat{SE}^\phi_S)^2
 =\hat p(1-\hat p)\,(1/\tilde n_S+1/\tilde n_{\bar S})\,(1+O(|\Delta|))$,
because $\hat p_S(1-\hat p_S)$ and $\hat p_{\bar S}(1-\hat p_{\bar S})$
differ from $\hat p(1-\hat p)$ by $O(|\Delta|)$.  Under the null and under local alternatives, where disparities are of
the order of their standard error, $\Delta\to0$ in probability; this
is the range in which the gate operates.  There
$\Lambda_S=Z_S^2\,(1+o_p(1))$, and for nonzero $\Delta$
Eq.~\eqref{eq:se-eff} gives
$\widehat{SE}_{\mathrm{eff},S}=\widehat{SE}^\phi_S\,(1+o_p(1))$.

\paragraph{Null law of the soft-count statistic.}
Fix $\tau>0$ and the predictor independently of the sample.
Assume independent observations within and across groups, identically
distributed within each group, with a common soft mean $p\in(0,1)$
under the null and $\tilde n_S/N\to\rho\in(0,1)$.  For fractional
counts the Bernoulli form overstates the variance: in each group
$\operatorname{Var}(Q_i)=p(1-p)-\mathbb{E}[Q_i(1-Q_i)]$, and the central
limit theorem applied to $\Delta$ in the expansion above gives
$\Lambda_S\Rightarrow c\,\chi^2_1$ with
\[
    0\le c=\frac{(1-\rho)\,v_S+\rho\,v_{\bar S}}{p(1-p)}\le 1,
    \qquad
    v_S=\operatorname{Var}(Q_i\mid i\in\Pi_S),\quad
    v_{\bar S}=\operatorname{Var}(Q_i\mid i\in\Pi_{\bar S}).
\]
The sampling standard error of the soft mean contrast is
$\sqrt{v_S/\tilde n_S+v_{\bar S}/\tilde n_{\bar S}}$; the Bernoulli-form
denominator is a proxy for it.  For $c>0$, the reference gate opens
under soft parity with asymptotic probability
$P(\chi^2_1>z_\alpha^2/c)\le\alpha$.  The same factor $c$ scales $Z_S^2$, preserving the local
equivalence of the two statistics.  For $c>0$, division by a consistent
estimate of $c$ would give asymptotic nominal size under these same
fixed-predictor, soft-null assumptions; SAFER retains the Bernoulli
reference cutoff.

For exact hard indicators $\mathbf1\{f_\theta(X)>1/2\}$, the
corresponding deviance instead has Wilks' $\chi^2_1$ null limit when
the common hard positive rate is interior.  To relate this to soft
counts as $\tau\downarrow0$, assume that neither conditional score
distribution has mass at $1/2$ and that the limiting hard rates are
interior.  Bounded convergence then gives
$\operatorname{Var}(Q_i)/[(\mathbb E Q_i)(1-\mathbb E Q_i)]\to1$
in each group, and hence $c\to1$ along models satisfying soft parity.

At fixed $\tau$, equal hard positive rates need not give equal
soft means, so the null of the gate is soft parity, and certification
of hard-decision parity is the task of the held-out hard-count audit
of Section~\ref{sec:measurement}.  The fixed-predictor result above
applies when the statistic is evaluated on data independent of the
fit.  During training the sample is also used to fit $f_\theta$, and
there the gate is an optimization device that uses the audit's
reference cutoff.

\subsection{Gradient identities for Section~\ref{sec:safer-densetail}}
\label{app:dense-tail}

Recall from Section~\ref{sec:safer-densetail} that
\(u_S(\theta) = \sqrt{\widehat\phi^\tau_S(\theta)^2+\varepsilon^2}
- z_\alpha\,\operatorname{sg}[\widehat{SE}_{\mathrm{eff},S}(\theta)]\),
where \(\operatorname{sg}[\cdot]\) denotes stop-gradient.  Since the
stop-gradient makes the threshold a constant for differentiation,
\(\nabla_\theta u_S =
\nabla_\theta\sqrt{\widehat\phi^\tau_S(\theta)^2+\varepsilon^2}\).
By the chain rule and
\(\operatorname{softplus}'_{\tau_s}(u)=\sigma(u/\tau_s)\),
\[
    \partial_\theta\operatorname{softplus}_{\tau_s}(u_S(\theta))
    =
    \sigma(u_S/\tau_s)\,\partial_\theta u_S(\theta).
\]
Taking norms gives Eq.~\eqref{eq:hinge-suppression}.  If \(u_S\le0\), then
\(\sigma(u_S/\tau_s)\le1/2\); if \(u_S<0\), then
\(\sigma(u_S/\tau_s)\le \exp(u_S/\tau_s)\).

The dense-tail contribution for subgroup \(S\) is
\(w_{\mathrm{tail}}\sqrt{\Lambda_S(\theta)}\).
Differentiating gives
\(w_{\mathrm{tail}}\partial_\theta\sqrt{\Lambda_S(\theta)}\).
This expression contains no factor \(\sigma(u_S/\tau_s)\), because the
LRT statistic is computed from the soft counts directly and does not
pass through the significant-excess softplus.

\paragraph{Size of the suppression.}
With \(\tau_s=10^{-2}\), the multiplier \(\sigma(u_S/\tau_s)\) equals
\(1/2\) at the gate boundary \(u_S=0\), \(0.0067\) at
\(u_S=-0.05\), and \(0.9933\) at \(u_S=0.05\).  The suppression below the
boundary is the intended safety mechanism, but it also acts on active
subgroups that strong regularization pushes back toward the boundary, so
the high-regularization frontier can stall; the dense tail supplies a weak
gradient path that does not contain this multiplier.

\paragraph{Why \(\sqrt{\Lambda_S}\).}
Three candidates have the required descent direction.  An ungated
raw-disparity term \(|\widehat\phi^\tau_S|\) would be certainty-blind and
would penalize noisy disparities regardless of their statistical support.
The statistic \(\Lambda_S\) itself would place the tail on a squared
evidence scale and produce much steeper precision-dependent gradients.
The intermediate choice \(\sqrt{\Lambda_S}\) remains on the same
standardized evidence scale as the gate and satisfies
\(\sqrt{\Lambda_S}=(|\widehat\phi^\tau_S|/\widehat{SE}^\phi_S)(1+o_p(1))\)
near the gate boundary (property (iii) in
Section~\ref{sec:safer-lrt}), so it retains the uncertainty weighting
without introducing a new estimator or threshold; its strength is governed
by the single weight \(w_{\mathrm{tail}}\).

\paragraph{Effect on the safety guarantee.}
Because the tail is ungated, Eq.~\eqref{eq:safety-tail-bound} carries an
additive residual proportional to \(w_{\mathrm{tail}}\), whose contribution
to the complete objective scales as \(\lambda w_{\mathrm{tail}}\); only the
gated component retains the exponentially small gradient guarantee below
the gate boundary.  Setting \(w_{\mathrm{tail}}=0\) removes the
residual and recovers the gated-only objective.  All full-SAFER runs use
the fixed value \(w_{\mathrm{tail}}=0.003\); the NoDenseTail ablation sets
it to zero.  The proposition bounds the back-propagated fairness
contribution at one point.  How far
training moves from the ERM optimum depends also on the curvature of the
prediction loss there: for a scalar quadratic prediction loss with curvature
\(\kappa\), a constant residual gradient of size \(r\) shifts the minimizer by
\(r/\kappa\).  The hypothesis \(u_S\le-\epsilon^\star\) is a margin
condition: an empty active set alone gives only the multiplier bound
\(\sigma(u_S/\tau_s)\le1/2\), and the exponential bound needs
\(\epsilon^\star\) large relative to \(\tau_s\).

\subsection{Operating regimes: the gate boundary}
\label{app:regimes}
The regimes describe the state of the gate at a checkpoint.  The
implemented boundary is \(u_S=0\); in the regular unclamped case
with nonzero disparity the active side \(u_S>0\) is \(T_S=\sqrt{\Lambda_S}>z_\alpha\).  Softplus
makes the transition between the gate states continuous through the
multiplier \(\sigma(u_S/\tau_s)\) of Eq.~\eqref{eq:hinge-suppression},
whose values at \(u_S=\pm0.05\) are given above and give a concrete
meaning to ``well below'' and ``well above'' the boundary.  There is no
universal sample-size boundary.  For EO the local approximation
\[
T_S\approx
\frac{|\phi_S|}
{\sqrt{p_S(1-p_S)/m_S+p_{\bar S}(1-p_{\bar S})/m_{\bar S}}}
\]
shows that certification depends jointly on the disparity magnitude, the
outcome rates, and the positive-label counts \(m_S\), \(m_{\bar S}\) on
both sides; for SP the total subgroup counts \(n_S\), \(n_{\bar S}\) take
their place.

\subsection{Stop-gradient and the threshold across updates}
\label{app:stop-gradient}
Write $d_S(\theta)=\sqrt{\widehat\phi_S^\tau(\theta)^2+\varepsilon^2}$
for the smoothed disparity and
$b_S(\theta)=z_\alpha\widehat{SE}_{\mathrm{eff},S}(\theta)$ for the
threshold.  At iteration $k$, SAFER differentiates
$\ell_S(\theta)=\operatorname{softplus}_{\tau_s}\bigl(d_S(\theta)-\operatorname{sg}[b_S(\theta)]\bigr)$,
so that
\[
\nabla_\theta\ell_S(\theta_k)=
\sigma\!\left(\frac{d_{S,k}-b_{S,k}}{\tau_s}\right)\nabla_\theta d_S(\theta_k).
\]
The threshold $b_{S,k}$ is held fixed during that backward pass only.
After the parameters are updated, $b_{S,k+1}=b_S(\theta_{k+1})$ is
recomputed from the new predictions and can increase or decrease.

\paragraph{A scalar example.}
Let $d(\theta)=\theta$, $b(\theta)=2\theta$ and $\theta_k=1$, and
absorb the positive softplus factor and the learning rate into an
effective step of $0.1$; the example sets the gate factor aside and
shows only the direction of the two gradient routes.  With stop-gradient, the direct term
differentiates only $d$, giving $\theta_{k+1}=0.9$; on the next forward
pass the threshold is recomputed and changes from $2$ to $1.8$.  Without
stop-gradient, the derivative of the excess is $d'(\theta)-b'(\theta)=-1$,
so descent instead gives $\theta_{k+1}=1.1$: the disparity increases from
$1$ to $1.1$, but the threshold rises faster, from $2$ to $2.2$, and the
excess decreases.  Detachment removes this route to a lower penalty.

\paragraph{Threshold drift in practice.}
We logged the unweighted mean of $b_S$ over the subgroups eligible
for the EO loss (at least two positive-label examples in both $S$ and
$\bar S$: 156 subgroups on Adult and 35 on COMPAS) together with the
complete fairness loss.  Table~\ref{tab:threshold-drift} reports means
$\pm$ standard deviations over 10 Adult and 5 COMPAS runs; ``final'' is
epoch 199 on Adult and each seed's early-stopped epoch on COMPAS.  Near
initialization the scores are approximately constant, so disparities and
the fairness loss are small, and both grow as the prediction loss learns
predictive structure.  The mean threshold moves from $0.281$ to $0.269$ on
Adult and from $0.128$ to $0.110$ on COMPAS.  These are averages over
subgroups, and individual thresholds can rise; the guarantee itself
concerns one update, in which the gated term supplies no gradient through
$b_S$.

\begin{table}[h]
\centering\small
\caption{Mean gate threshold $b_S$ over eligible subgroups
and complete fairness loss at the start and end of training (mean $\pm$
standard deviation over runs).}
\label{tab:threshold-drift}
\begin{tabular}{lcc}
\toprule
dataset (runs) & mean $b_S$: initial / final & fairness loss: initial / final \\
\midrule
Adult (10) & $.281\pm.032$ / $.269\pm.014$ & $.0015\pm.0007$ / $.0037\pm.0011$ \\
COMPAS (5) & $.128\pm.021$ / $.110\pm.003$ & $.0034\pm.0045$ / $.0205\pm.0023$ \\
\bottomrule
\end{tabular}
\end{table}

\subsection{Proof of Proposition~\ref{cor:safety}}

At \(\theta^\star_{\mathrm{ERM}}\), assume
\(u_S(\theta^\star_{\mathrm{ERM}})\le-\epsilon^\star\) for all \(S\).  By
Eq.~\eqref{eq:hinge-suppression},
\[
    \big\|
    \partial_\theta
    \operatorname{softplus}_{\tau_s}(u_S(\theta^\star_{\mathrm{ERM}}))
    \big\|
    \le
    e^{-\epsilon^\star/\tau_s}
    \big\|\partial_\theta u_S(\theta^\star_{\mathrm{ERM}})\big\|.
\]
The bounded-gradient regularity condition in Proposition~\ref{cor:safety}
allows the finite average of \(\|\partial_\theta u_S\|\) to be absorbed into
the constant.  Summing over subgroups gives
Eq.~\eqref{eq:safety-bound}.  Adding the dense-tail gradients and applying
the triangle inequality gives Eq.~\eqref{eq:safety-tail-bound}. \(\square\)

\end{document}